\documentclass[sigconf]{acmart}
\AtBeginDocument{%
  }

\usepackage{amsmath}
\usepackage{mathtools}
\usepackage{amsthm}

\usepackage{algorithmic}
\usepackage{algorithm}
\usepackage{booktabs}
\usepackage{multirow}
\usepackage{makecell}
\usepackage{subcaption}
\usepackage[table]{xcolor}
\usepackage{colortbl}

\theoremstyle{plain}

\newtheorem{lemma}{Lemma}

\theoremstyle{definition}
\newtheorem{definition}{Definition}[section]
\newtheorem{theorem}{Theorem}[section]

\theoremstyle{remark}

\setcopyright{cc}
\setcctype{by}
\copyrightyear{2027}
\acmYear{2027}
\acmDOI{10.1145/3862868.3871127}
\acmConference[KDD 2027] {Proceedings of the 33rd ACM SIGKDD Conference on Knowledge Discovery and Data Mining V.1}{August 1--5, 2027}{San Jose, CA, USA.}
\acmBooktitle{Proceedings of the 33rd ACM SIGKDD Conference on Knowledge Discovery and Data Mining V.1 (KDD 2027), August 1--5, 2027, San Jose, CA, USA}
\acmISBN{979-8-4007-3134-1/2027/08}

\begin{document}

\title{Hypergraph Structural Entropy-Enhanced Hierarchy-Aware Semi-Supervised Graph Classification}

\author{Guangjie Zeng}
\authornote{Guangjie Zeng and Hao Peng are corresponding authors.}
\affiliation{%
  \institution{Northeastern University at Qinhuangdao
  }
  \department{Hebei Key Laboratory of Marine Perception Network and Data Processing}
  \city{Qinhuangdao}
  \state{Hebei}
  \country{China}}
\email{zengguangjie@neuq.edu.cn}

\author{Hao Peng}
\authornotemark[2]
\affiliation{%
  \institution{Beihang University}
  \department{School of Cyber Science and Technology}
  \city{Beijing}
  \country{China}}
\email{penghao@buaa.edu.cn}

\author{Angsheng Li}
\affiliation{%
  \institution{Beihang University}
  \department{School of Computer Science}
  \city{Beijing}
  \country{China}}
\email{angsheng@buaa.edu.cn}

\author{Li Sun}
\affiliation{%
  \institution{Beijing University of Post and Telecommunications}
  \city{Beijing}
  \country{China}
}
\email{lsun@bupt.edu.cn}

\author{Chunyang Liu}
\affiliation{%
 \institution{Didi Chuxing}
 \city{Beijing}
 \country{China}}
\email{liuchunyang@didiglobal.com}

\author{Yicheng Pan}
\affiliation{%
  \institution{Beihang University}
  \city{Beijing}
  \country{China}}
\email{yichengp@buaa.edu.cn}

\author{Shengze Li}
\affiliation{%
  \institution{Academy of Military Science}
  \city{Beijing}
  \country{China}}
\email{lsz86591989@163.com}

\author{Philip S. Yu}
\affiliation{%
  \institution{University of Illinois Chicago}
  \city{Chicago}
  \state{Illinois}
  \country{USA}}
\email{psyu@uic.edu}

\renewcommand{\shortauthors}{Guangjie Zeng et al.}

\begin{abstract}
Semi-supervised graph classification aims to learn graph-level representations by leveraging a large number of unlabeled graphs alongside a small set of labeled ones, effectively mitigating the issue of label scarcity.
Despite significant progress by various semi-supervised learning strategies, complex relationships among graph vertices beyond pairwise edges, especially hierarchical structures, remain underexplored by existing methods. 
In this work, we propose a \textbf{Hyp}ergraph \textbf{S}tructural \textbf{E}ntropy-\textbf{E}nhanced hierarchy-aware framework, namely HypSEE, for semi-supervised graph classification.
Specifically, we first construct walk-induced and embedding-derived hypergraphs to capture hierarchy-aware local and long-range higher-order relationships within graphs.
Then, we devise a hierarchical hypergraph pooling approach guided by hypergraph structural entropy (HSE) to facilitate hierarchical representation learning in hypergraphs.
Lastly, we employ an anchor graph-based consistency strategy to mutually align the representations from these two hypergraph views, enhancing the learning efficacy on unlabeled data.
Extensive experiments on 8 benchmark datasets demonstrate the state-of-the-art performance of our approach.
\end{abstract}




\maketitle

\section{Introduction}
Semi-supervised graph classification (SSGC) focuses on predicting the graph classes when only a limited number of labels are available \cite{luo2022dualgraph}.
Unlike supervised graph classification, SSGC methods leverage a large number of unlabeled graphs under the guidance of a small set of labeled graphs.
These approaches enable effective learning from unlabeled graphs, eliminating the need for large-scale graph annotation and reducing the burden on domain experts.
SSGC is particularly valuable for domains where graph data is abundant, but graph annotation is scarce and costly, such as molecular property prediction \cite{hao2020asgn}, social group classification \cite{li2019semi}, and network-based research paper classification \cite{li2022semi}.

Driven by their powerful expressive capabilities, Graph Neural Networks (GNNs) \cite{kipf2017semi} currently dominate SSGC. 
To effectively harness unlabeled graph data, pseudo-labeling approaches \cite{luo2022dualgraph} iteratively expand the limited labeled set by leveraging high-confidence model predictions. 
In parallel, consistency learning methods \cite{suninfograph} extract self-supervised signals by aligning representations across diverse structural or feature views. 
Building upon these foundational paradigms, subsequent research has introduced various enhancements, including graph kernel-based encoders for better topological modeling \cite{ju2022tgnn,ju2022kgnn}, advanced graph augmentations \cite{you2020graph,yue2022label}, and representation disentanglement for robust knowledge transfer \cite{wang2025disensemi}.

Despite the abundance and promising performance of existing GNN methods, most of them overlook complex relationships among graph vertices beyond single edges.
Traditional GNNs \cite{kipf2017semi} aggregate information from the direct neighborhood of vertices at each layer.
Stacking multiple GNN layers to reach vertices at a long range incurs the over-smoothing problem.
Moreover, GNNs struggle to model the relationships among multiple closely related vertices such as aromatic functional groups in a molecule, which are important in determining graph properties.
While methods like HEAL \cite{ju2024hypergraph} employ hypergraphs to model higher-order dependencies among vertices, they do not adequately explore the hierarchical structures of graphs.
Real-world graphs, such as social networks, may exhibit insightful hierarchical community structures, which are particularly relevant to graph classification.
For example, the properties of an organic molecule are determined by both single atoms and aromatic functional groups.
Neglecting these hierarchy structures may degrade the overall prediction performance.
Therefore, a primary challenge of SSGC is the under-exploration of higher-order relationships, particularly hierarchical structures in graphs.

\begin{figure}[t]
\centering
\includegraphics[width=0.9\linewidth]{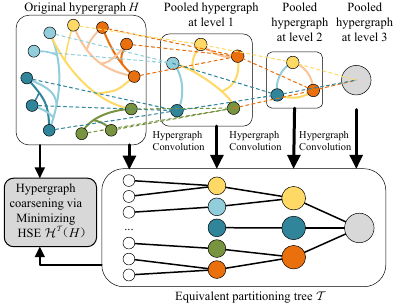}
\caption{Hypergraph coarsening via HSE minimization.
}
\Description{A schematic illustration of hypergraph coarsening via HSE minimization.}
\label{fig_demo}
\end{figure}

To address this challenge, we propose a \textbf{Hyp}ergraph \textbf{S}tructural \textbf{E}ntropy-\textbf{E}nhanced hierarchy-aware framework, namely HypSEE, for SSGC.
This framework employs hypergraphs to model long-range dependencies and complex multi-way relationships among vertices, and these hypergraphs are coarsened into various sizes by HSE-guided pooling to capture hierarchical structures.
Specifically, we construct two complementary hypergraph views using distinct strategies to comprehensively explore higher-order node interactions.
The first, a random walk-induced view, employs random walks of varying lengths to capture local, hierarchy-aware dependencies.
The second, an embedding-derived view, dynamically constructs flexible hyperedges parameterized by GNNs to model global, long-range relationships.
To facilitate hierarchical representation learning, we stack hypergraph convolution and pooling layers, as illustrated in Fig. \ref{fig_demo}.
Crucially, the hypergraph coarsening process is guided by the minimization of Hypergraph Structural Entropy (HSE). 
While classic structural entropy (SE) \cite{li2016structural,li2024science} is proven effective for standard graphs, we extend this principle to HSE to explicitly decode the hierarchical organizations within hypergraph topologies.
Finally, we introduce an anchor graph-based consistency learning strategy across the two views. By sampling a subset of labeled graphs as anchors, we enforce representation alignment between the dual views and these anchors.
This consistency regularization effectively leverages unlabeled data to provide robust self-supervisory signals, thereby aligning the semantic spaces of different views and alleviating the scarcity of labeled data in SSGC.

The main contributions are summarized as follows:
(1) We extend HSE from classic SE to explore the hierarchy structure learning in hypergraphs.
(2) We devise a hypergraph hierarchical representation learning approach by HSE-guided hierarchical graph pooling, which captures the hierarchy-aware higher-order relations within graphs to enhance representation ability.
(3) We propose HypSEE, a novel SSGC framework that enhances classification accuracy by forcing consistency between two hypergraph views from graphs.
Extensive experimental results on 8 datasets demonstrate the state-of-the-art performance of HypSEE.

\section{Preliminaries}
\subsection{Hypergraph}
\begin{definition}
[Hypergraph \cite{zhou2006learning}]\label{def_hypergraph}
An attributed hypergraph can be defined as a $H = (\mathcal{V},\mathcal{E}_H, \mathbf{X}, \mathbf{W})$ with the vertex set $\mathcal{V}$, the hyperedge set $\mathcal{E}_H$, the vertex feature matrix $\mathbf{X}$, and the diagonal hyperedge weight matrix $\mathbf{W}\in \mathbb{R}^{|\mathcal{E}_H| \times |\mathcal{E}_H|}$, where each hyperedge $e_H \in \mathcal{E}_H$ is a subset of $\mathcal{V}$ and can connect more than two vertices.
The hypergraph $H$ can also be described by an incidence matrix $\mathbf{H} \in \mathbb{R}^{|\mathcal{V}|\times|\mathcal{E}_H|}$, 
where $\mathbf{H}(v,e_H) = 1$ if $v \in e_H$ and $0$ otherwise.
\end{definition}

\noindent\textbf{Hypergraph Cut \& Volume.}
Given a hypergraph $H = (\mathcal{V}, \mathcal{E}_H, \mathbf{W})$ and a vertex subset $S \subset \mathcal{V}$, the complement $S^c$ of $S$ is defined as $S^c = \{v|v\notin S, v\in \mathcal{V}\}$, and the hyperedge boundary $\partial S$ is defined as $\partial S = \{ e_H \in \mathcal{E}_H | e_H \cap S \neq \emptyset, e_H \cap S^c \neq \emptyset \}$ containing hyperedges that connect to both vertices in $S$ and $S^c$.
The cut $g(S)$ of $S$, i.e., the volume $V_{\partial S}$ of $\partial S$, is a penalty of partitioning $\mathcal{V}$ into $S$ and $S^c$, while the volume $V_S$ of $S$ is the sum of degrees of vertices in $S$ \cite{zhou2006learning}.
Formally, $V_S$ is defined as $V_S = \sum_{v \in S} d_v$, where $d_v$ is the degree of vertex $v$ defined as $d_v = \sum_{\{e_H \in \mathcal{E} | v\in e_H \}} \mathbf{W}(e_H)$.
The cut $g(S)$ is defined as follows:
\begin{equation}
g(S) = V_{\partial S} = \sum_{e_H \in \partial S} \mathbf{W}(e_H)\frac{|e_H\cap S| |e_H\cap S^c|}{|e_H|},
\end{equation}
where $|e_H|$ is the cardinality of the set corresponding to $e_H$.

\subsection{Structural Entropy}
SE \cite{li2016structural} of a graph $G$ with an associated encoding tree $\mathcal{T}$ quantifies the uncertainty in $G$ after decoding by $\mathcal{T}$.
The optimal encoding tree $\mathcal{T}^*$ of $G$ is the one with the minimum SE that best reveals the hierarchical structure of vertices in $G$ in the context of SE.
As a versatile graph metric, SE has demonstrated success in various domains, including social event detection \cite{liu2026effective}, spatial cell clustering  \cite{huang2026sector}, 3D genome analysis \cite{li2023dedoc2}, and offline reinforcement learning \cite{zeng2026structural}.
Due to the intrinsic hierarchical nature of encoding trees, SE is particularly effective at capturing multi-level structures, such as hierarchical clustering \cite{zeng2026hyperbolic}, hierarchical graph pooling \cite{wu2022structural}, and hierarchical text classification \cite{zhu2023hitin}.
We give the definition of classic SE \cite{li2016structural}, and then introduce the level-wise assignment \cite{sunlsenet} that HSE is based on.

\begin{definition}
[Structural Entropy (\cite{li2016structural})]\label{def:se}
Given an undirected weighted graph $G = (\mathcal{V},\mathcal{E})$ and an associated encoding tree $\mathcal{T}$ with root $\lambda$, the SE of $G$ on $\mathcal{T}$ is defined as
\begin{equation}\label{eq:se}
    \mathcal{H}^\mathcal{T}(G) = \sum_{\alpha\in\mathcal{T},\alpha\neq\lambda}\mathcal{H}^\mathcal{T}(G;\alpha)
    = \sum_{\alpha\in\mathcal{T},\alpha\neq\lambda} - \frac{g_\alpha}{V_G}\log_2\frac{V_\alpha}{V_{\alpha^-}},
\end{equation}
where $\mathcal{H}^\mathcal{T}(G; \alpha)$ is the SE assigned to a node $\alpha$ in $\mathcal{T}$.
Each node $\alpha$ in $\mathcal{T}$ is associated with a vertex set $T_\alpha$, where $T_\lambda = \mathcal{V}$ and $T_\nu$ of leaf node $\nu$ contains only one vertex in $\mathcal{V}$.
For each non-leaf node $\alpha$ in $\mathcal{T}$, its children are associated with disjoint vertex subsets, whose union is $T_\alpha$.
The parent of non-root node $\alpha$ is denoted as $\alpha^-$.
The notation $g_{\alpha}$ is the cut, i.e., the sum of graph edge weights with exactly one endpoint in $T_{\alpha}$.
Notations $V_{\alpha}$ and $V_{G}$ are the volumes, i.e., the sum of node degrees in $T_{\alpha}$ and $T_G$, respectively.
To meet the requirements of downstream applications, the $K$-dimensional SE is defined as the SE of $G$ on any encoding tree with a depth of $K$.
\end{definition}

\begin{definition}
[Level-Wise Assignment \cite{sunlsenet}]\label{def:assignment}
Given a partitioning tree $\mathcal{T}$ with depth $K$, a list of level-wise assignment matrices $\{ \mathbf{C}^h \}_{h=1}^{K+1}$ contains all parent relationships in $\mathcal{T}$.
The matrix $\mathbf{C}^h \in \{ 0,1 \}^{N_h\times N_{h-1}}$ represents the parent assignment of tree nodes from $h$-th to $h-1$-th level, where $\mathbf{C}^h_{ik} = 1$ means the $k$-th node in $\mathcal{T}$ at depth $h-1$ is the parent node of $i$-th node at depth $h$.
With $\{ \mathbf{C}^h \}_{h=1}^{K+1}$, we can derive node assignment matrices $\mathbf{S^h}$ at all depths representing the ascendant-descendant relationship between tree nodes at depth $h$ and graph vertices, i.e., tree leaves.
Specifically, $\mathbf{S}^h_{ik} = 1$ means the $k$-th node in $\mathcal{T}$ at depth $h$ is an ascendant of the $i$-th leaf, i.e., the $i$-th vertex in $\mathcal{V}$ belongs to associated vertex set $T_{\alpha_k^h}$ of $k$-th node in $\mathcal{T}$ at depth $h$.
The matrix $\mathbf{S}$ can be derived by
\begin{equation}\label{eq:S}
\mathbf{S}^h = \prod\nolimits_{k=K+1}^{h+1}\mathbf{C}^k,  \quad \mathbf{C}^{K+1}=\mathbf{I}_{|\mathcal{V}|},
\end{equation}
where $\mathbf{I}_{|\mathcal{V}|}$ is an identity matrix.
\end{definition}

\section{Methodology}

\begin{figure*}[t]
\centering
\includegraphics[width=0.9\linewidth]{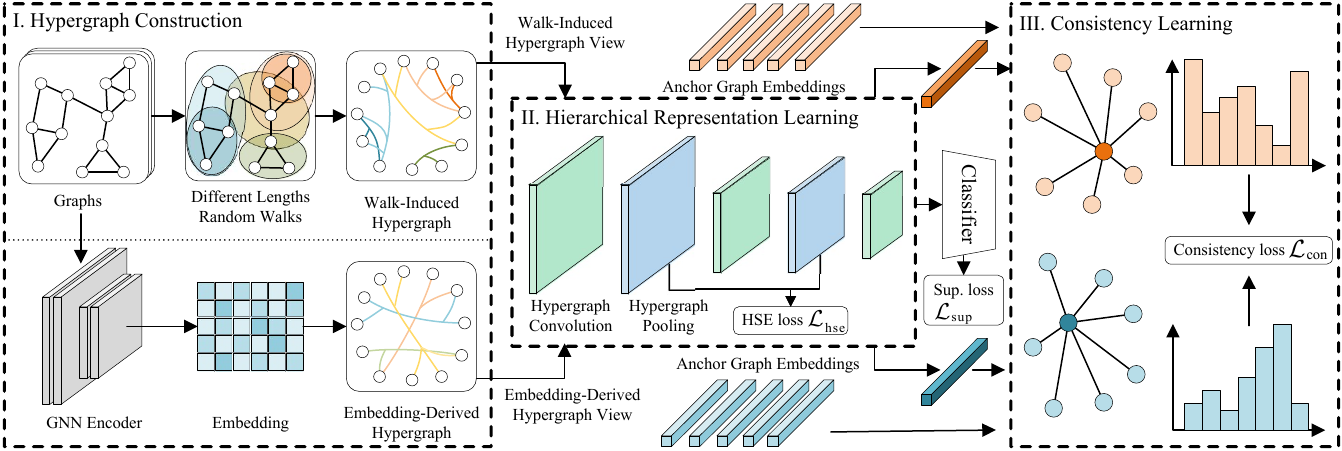}
\caption{Framework of HypSEE.
(I) The Hypergraph Construction module generates the walk-induced and the embedding-derived hypergraph view from graphs.
(II) The Hierarchical Representation Learning module encodes hypergraphs from both views.
(III) The Consistency Learning module enhances representation quality by forcing consistency of two different views.
}
\Description{Framework of HypSEE.}
\label{fig_framework}
\end{figure*}

In SSGC, label prediction is performed when a large number of graphs are available but only a few are labeled.
Let $G = (\mathcal{V}, \mathcal{E}, \mathbf{X})$ denotes an attributed undirected graph with the vertex set $\mathcal{V}$, the edge set $\mathcal{E}$, and the feature matrix $\mathbf{X}$, we aim to predict the class label $\mathbf{y}$ of $G$ if not given.
Given a set of labeled graphs $\mathcal{G}^L = \{ G_1,\ldots,G_{| \mathcal{G}_L |} \}$ with labels $\mathcal{Y}^L = \{ \mathbf{y}_1,\ldots,\mathbf{y}_{| \mathcal{G}_L |} \}$, and a set of unlabeled graphs $\mathcal{G}^U = \{ G_{| \mathcal{G}^L+1 |},\ldots, G_{| \mathcal{G}^L | + | \mathcal{G}^U |} \}$, SSGC aims to predict the set of labels $\mathcal{Y}^U$ of the unlabeled graphs in $\mathcal{G}^U$, where $| \mathcal{G}^U | \gg | \mathcal{G}^L |$ in most cases.
As illustrated in Fig. \ref{fig_framework}, HypSEE first constructs two different hypergraph views from graphs, then encodes hypergraphs via \textit{Hierarchical Representation Learning}, and enhances representation quality by \textit{Consistency Learning}.

\subsection{Hypergraph Construction}
To model complex interactions beyond standard pairwise edges, we transform the input graphs into hypergraphs.
Specifically, we propose a dual-view construction strategy: the first view captures hierarchy-aware local structures, while the second models long-range topological dependencies.
These two complementary representations synergistically enhance the learning process, establishing a robust foundation for exploiting unlabeled data through our subsequent anchor graph-based consistency learning.

\noindent\textbf{Walk-Induced Hypergraph View.}
Higher-order relationships among graph vertices can exhibit different lengths reflecting substructures in graphs of different sizes.
We construct a hierarchy-aware hypergraph by random walk \cite{grover2016node2vec} at different lengths from each graph to better capture the hierarchy structure in the graph.
For a vertex $u$ in graph $G=(\mathcal{V},\mathcal{E})$, we simulate a random walk starting from $u$ by generating a sequence of vertices according to probability as follows:
\begin{equation}
P(c_{i+1}=x \ | \  c_i = v) =
\begin{cases}
\frac{1}{\pi} & \text{if } (v, x) \in \mathcal{E}, \\
0 & \text{otherwise}.
\end{cases}
\end{equation}
where $\pi$ is the number of candidate vertices.
Given $t$, the last traversed vertex, candidate vertex $x$ is selected from the vertices connecting to $v$ and within a distance less than 2 from $t$.
We set $c_0=u$ and the generated sequence is then $c_0,c_1,\ldots,c_{l_\epsilon}$ functioning as a length $l_\epsilon$ hyperedge $e_H$.
To fully capture the higher-order relations at different lengths around different vertices, we randomly sample $K^w$ starting vertices from $\mathcal{V}$ and set the random walk lengths from a predefined hyperedge length set $\{l_\epsilon\}$.
These hyperedges form the walk-induced hypergraph $H^w$ with the incidence matrix $\mathbf{H}^w$.

\noindent\textbf{Embedding-Derived Hypergraph View.}
Hypergraphs constructed using hand-crafted criteria, such as random walks, may result in sub-optimal performance due to their limited flexibility.
To address this issue, we turn to trainable hypergraphs parameterized by GNNs \cite{sun2023self,ju2024hypergraph}, which are jointly optimized with the downstream graph classification.
Given a graph $G=(\mathcal{V},\mathcal{E})$ with affinity $\mathbf{A}$ and feature $\mathbf{X}$, we derive graph embedding $\mathbf{X}^e$ by a three-layer GCN \cite{kipf2017semi} as:
\begin{equation}
\mathbf{X}^{(l+1)} = \mathrm{ReLU}(\tilde{\mathbf{D}}^{-\frac{1}{2}} \tilde{\mathbf{A}} \tilde{\mathbf{D}}^{-\frac{1}{2}} \mathbf{X}^{(l)}\mathbf{\Psi}^{(l)}), \ \ \tilde{\mathbf{A}} = \mathbf{A} + \mathbf{I}_{|\mathcal{V}|},
\end{equation}
where $\mathbf{X}^{(l)}$ and $\mathbf{\Psi}^{(l)}$ are the graph embedding and learnable weights at $l$ layer, $\mathbf{X}^{(0)}=\mathbf{X}$, $\tilde{\mathbf{D}}_{ii} = \Sigma_j \tilde{\mathbf{A}}_{ij}$, $\mathbf{I}_{|\mathcal{V}|}$ is the identity matrix, and $\mathbf{X}^e = \mathbf{X}^{(3)} \in \mathbb{R}^{|\mathcal{V}| \times d^e}$ is the derived embedding with $d^e$ dimension. We then derive the learnable hypergraph $H^e$ with the incidence matrix $\mathbf{H}^e \in \mathbb{R}^{|\mathcal{V}| \times K^e}$ containing $K^e$ hyperedges as follows:
\begin{equation}
\mathbf{H}^e = \mathrm{Sigmoid}(\mathbf{X}^e \mathbf{\Psi}^e),
\end{equation}
where $\mathbf{\Psi}^e \in \mathbb{R}^{d^e \times K^e}$ is a learnable weight matrix and the $\mathrm{Sigmoid}$ activation is used to generate meaningful hyperedges.

\subsection{Hierarchical Representation Learning}
Hierarchical organization is a fundamental topological property that encodes valuable relative positions and influence patterns among vertices.
Enabling multi-level information propagation within GNNs is therefore crucial for comprehensively capturing these structures to encode the entire graph.
While abundant pooling methods exist for supervised graph classification \cite{ying2018hierarchical,bianchi2020spectral}, this structural prior is largely overlooked by current SSGC methods.
Furthermore, existing pooling mechanisms fail to accommodate hypergraphs, restricting their capacity to learn higher-order relationships.
To bridge these gaps, we propose a novel pooling approach guided by HSE.
Traditional SE \cite{li2016structural} naturally quantifies a graph's topological complexity via a partitioning tree, making it an ideal tool for extracting hierarchical structures \cite{wu2022structural,ren2024hi}.
By generalizing classic SE to the hypergraph domain, we establish a theoretically grounded criterion for hierarchical hypergraph pooling, detailed as follows.
The details of HSE and hypergraph hierarchical pooling are presented below.

\noindent\textbf{Hypergraph Structural Entropy.}
To begin with, we introduce HSE to guide the hypergraph hierarchical pooling layers.
Consider a hypergraph $H = (\mathcal{V}, \mathcal{E}_H, \mathbf{W})$ and its corresponding incidence matrix $\mathbf{H}$.
We define the vertex degree matrix $\mathbf{D} \in \mathbb{R}^{|\mathcal{V}| \times |\mathcal{V}|}$ as a diagonal matrix, where the $i$-th entry $\mathbf{D}_{ii}$ represents the degree of vertex, defined as $\mathbf{D}_{ii}= \sum_{\epsilon=1}^{|\mathcal{E}_H|} \mathbf{W}_{\epsilon \epsilon} \mathbf{H}_{i \epsilon}$.
Analogously, the hyperedge degree matrix $\mathbf{B} \in \mathbb{R}^{|\mathcal{E}_H| \times |\mathcal{E}_H|}$ is a diagonal matrix where $\mathbf{B}_{\epsilon \epsilon}$ denotes the degree of hyperedge $\epsilon$, defined as $\mathbf{B}_{\epsilon \epsilon} = \sum_{i=1}^{|\mathcal{V}|}\mathbf{H}_{i \epsilon}$.
Given a sequence of level-wise assignment matrices $\{ \mathbf{C}^h \}_{h=1}^{K+1}$, we can iteratively derive the node assignment matrix $\mathbf{S}^h \in \mathbb{R}^{N \times N_h}$ at depth $h$.
The hypergraph volumes of the $k$-th tree node at depth $h$ and its parent node $k^-$ at depth $h-1$ are formulated as follows:
\begin{equation}
\mathbf{V}_k^h = \sum\nolimits_{i=1}^N \mathbf{S}_{ik}^h \mathbf{D}_{ii}, \ \ \ \ 
\mathbf{V}^{h-1}_{k^-} = \sum\nolimits_{k'=1}^{N_{h-1}}\mathbf{C}_{kk^\prime}^h \mathbf{V}^{h-1}_{k'}.
\end{equation}
The hypergraph cut of tree node $k$ at depth $h$ is given by:
\begin{equation}
\mathbf{M}_k^h = \sum\nolimits_{\epsilon=1}^{|\mathcal{E}_H|} 
\left[ \frac{\mathbf{W}_{\epsilon \epsilon}}{\mathbf{B}_{\epsilon \epsilon}}
   \sum\nolimits_{i=1}^{N}  \mathbf{S}_{ik}^h \mathbf{H}_{i\epsilon} 
    \sum\nolimits_{j=1}^N  \mathbf{H}_{j\epsilon} (1-\mathbf{S}_{jk}^h) \right].
\end{equation}
Taken together, we can derive the definition of $K$-dimensional HSE as follows.
\begin{definition}\label{def:hse}
[$K$-Dimensional Hypergraph Structural Entropy]
Given a hypergraph $H$ and an associated partitioning tree $\mathcal{T}$ with depth $K$, the HSE of $H$ w.r.t. $\mathcal{T}$ is defined as:
\begin{equation} \label{eq:hse}
\mathcal{H}^\mathcal{T}(H) = \sum_{h=1}^K \mathcal{H}^\mathcal{T}(H;h) = \sum_{h=1}^K - \frac{1}{V_H}  \sum_{k=1}^{N_h} \mathbf{M}_k^h \log_2 \frac{\mathbf{V}_k^h}{\mathbf{V}_{k^-}^{h-1}},
\end{equation}
where $\mathcal{H}^\mathcal{T}(H;h)$ is the HSE at depth $h$, $V_H = \sum\nolimits_{i=1}^N \mathbf{D}_{ii}$ is the volume of hypergraph $H$.
\end{definition}

\begin{theorem}[Connection to Classic Structural Entropy-Appendix \ref{sec:theorem1}]\label{theorem:degenerate}
The HSE of a simple graph $G$ w.r.t. any given partitioning tree $\mathcal{T}$ equals to $\frac{1}{2} \mathcal{H}^\mathcal{T}(G)$ in Equation (\ref{eq:se}).
\end{theorem}

The proof of Theorem \ref{theorem:degenerate} is provided in Appendix \ref{sec:theorem1}.
Theorem \ref{theorem:degenerate} reveals that HSE defined in Equation (\ref{eq:hse}) is a natural extension of the classic SE with a close relation.

\begin{theorem}[Connection to Hypergraph Clustering-Appendix \ref{sec:theorem2}]\label{theorem:clustering}
Given a hypergraph $H$ along with its incidence matrix $\mathbf{H}$, $\rho(H) = \mathcal{H}^K(H) / \mathcal{H}^1(H)$ is the normalized $K$-dimentional HSE of $H$, and $\Phi(H) = \min_{S \subseteq \mathcal{V}} \{ \frac{V_{\partial S}}{\min \{ V_{S}, V_{S^c} \}}\}$ is the hypergraph conductance. The following inequality holds:
\begin{equation}
\rho^K(H) \geq \Phi(H).
\end{equation}
\end{theorem}

The proof of Theorem \ref{theorem:clustering} is provided in Appendix \ref{sec:theorem2}.
Similar to graph conductance, hypergraph conductance is a widely used objective with different hypergraph cuts for hypergraph clustering \cite{takai2020hypergraph}.
Since the one-dimensional HSE $\mathcal{H}^1(H)$ is a constant for a given $H$, and it serves as an upper bound for hypergraph conductance (Theorem \ref{theorem:clustering}), minimizing HSE in Definition \ref{def:hse} is likely to also reduce hypergraph conductance.
Thus, HSE is capable of serving as a hypergraph clustering objective and guiding hypergraph coarsening for hierarchical representation learning.

\noindent\textbf{Hypergraph Hierarchical Pooling via HSE.}
The hypergraph hierarchical representation learning module stacks the hypergraph convolution layer \cite{gao2022hgnn+} and the hypergraph pooling layer alternately.
Through the hypergraph convolution layer, the vertex embeddings are aggregated into hyperedge embeddings and then transformed into the updated vertex embeddings.
Through the hypergraph pooling layer, an approximate assignment matrix $\mathbf{C}^h \in \mathbb{R}^{N_h\times N_{h-1}}$ at the current depth $h$ is learned for hypergraph coarsening.
The hierarchical representation of hypergraphs is effectively learned by performing feature aggregation in hypergraphs of different coarsening sizes.
Grouping these learned assignment matrices $\{ \mathbf{C}^h \}_{h=1}^{K+1}$ together, we obtain the approximate partitioning tree with a theoretical approximation guarantee \cite{sunlsenet}.
These hierarchical representations improve HypSEE's ability to capture hierarchical structures in hypergraphs.
For convenience, we denote the first hypergraph convolution and pooling layer to be at the depth $h=K$, and the following layers at $h=K-1$ to $h=1$.

For the hypergraph at depth $h$ represented by the incidence matrix $\mathbf{H}^h \in \mathbb{R}^{N_h \times |\mathcal{E}_H|}$ with vertex embeddings $\mathbf{X}^h \in \mathbb{R}^{N_h\times d}$, the hypergraph convolution layer computes the updated vertex embeddings $\mathbf{Z}^{h}$ as follows:
\begin{equation}\label{eq_hconv}
\mathbf{Z}^h = \sigma(\mathbf{D}^{-1}\mathbf{H}\mathbf{W}\mathbf{B}^{-1}\mathbf{H}^\top\mathbf{X}^h\mathbf{\Theta}^h)
\end{equation}
where $\mathbf{D}$ and $\mathbf{B}$ are weight matrices of the vertex and hyperedge, $\sigma(\cdot)$ is the activation function, and $\mathbf{\Theta}^h$ is the training weight matrix for convolution at depth $h$.
The subsequent pooling layer computes the assignment matrix $\mathbf{C}^h \in \mathbb{R}^{N_h \times N_{h-1}}$ by employing a differentiable $k$-means approach \cite{wilder2019end}. 
Specifically, we introduce a trainable matrix of cluster centers $\mathbf{U}^h \in \mathbb{R}^{N_{h-1} \times d}$ for depth $h$. 
The assignment probability of vertex $i$ to cluster $j$ is calculated by applying a Softmax function over the negative Euclidean distances between the vertex embeddings and the cluster centers:
\begin{equation}\label{eq_assignment}
\mathbf{C}^h_{ij} = \frac{\exp\left(-\|\mathbf{z}^h_i - \mathbf{u}^h_j\|^2 / \tau \right)}{\sum_{k=1}^{N_{h-1}} \exp\left(-\|\mathbf{z}^h_i - \mathbf{u}^h_k\|^2 / \tau \right)},
\end{equation}
where $\mathbf{z}^h_i$ and $\mathbf{u}^h_j$ are the $i$-th and $j$-th row vectors of $\mathbf{Z}^h$ and $\mathbf{U}^h$ respectively, and $\tau$ is a temperature parameter set as $\tau = 1$.
With the assignment matrix $\mathbf{C}^h$, we generate a coarsened hypergraph $\mathbf{H}^{h-1} \in \mathbb{R}^{N_{h-1} \times |\mathcal{E}_H|}$ and a new vertex embedding matrix $\mathbf{X}^{h-1} \in \mathbb{R}^{N_{h-1} \times d}$ as:
\begin{equation}
\mathbf{H}^{h-1} = (\mathbf{C}^h)^\top \mathbf{H}^h, \ \ \mathbf{X}^{h-1} = (\mathbf{C}^h)^\top \mathbf{Z}^h.
\end{equation}
These generated matrices serve as inputs for the next hypergraph convolution layer.

We stack one convolution layer followed by one pooling layer at depth $h>1$ and stack only one convolution layer at depth $h=1$ since the hypergraph $\mathbf{H}^1$ should be coarsened into the root of the partitioning tree.
The corresponding assignment matrix $\mathbf{C}^1 \in \mathbb{R}^{N_1 \times 1}$ is not trainable and set to $\mathbf{C}^1 = \mathbf{1}_{N_1 \times 1}$ full of $1$s.
The entire graph embedding is then calculated as $\mathbf{Z}^G = \mathbf{X}^0 = (\mathbf{C}^1)^ {\top} \mathbf{Z}^1$.
With all these assignment matrices, we can calculate the HSE as a loss function $\mathcal{L}_{\mathrm{hse}}$ by Eq. (\ref{eq:hse}) as follows:
\begin{equation}\label{eq_lhse}
\mathcal{L}_{\mathrm{hse}} (\mathbf{H}^K, \{\mathbf{C}^h\}_{h=1}^{K+1}) = \sum_{h=1}^K - \frac{1}{V_H}  \sum_{k=1}^{N_h} \mathbf{M}_k^h \log_2 \frac{\mathbf{V}_k^h}{\mathbf{V}_{k^-}^{h-1}}.
\end{equation}
This loss function is then jointly optimized with the SSGC task to regularize the graph coarsening procedures.

\subsection{Consistency Learning}
In the semi-supervised learning setting, the given labeled graphs of a small quantity are insufficient for model training, and the key to accurate graph classification is the effective utilization of unlabeled graphs of a large quantity.
The walk-induced hypergraph view and the embedding-derived hypergraph view in the hypergraph construction module extract the hierarchy-aware local and long-range relationships from graphs, respectively.
The representations from these two views complement each other, enabling effective learning from unlabeled graphs through consistency learning.
We adopt the anchor graph-based consistency learning strategy \cite{ju2022tgnn} to enhance the quality of graph representation.

Specifically, we sample a subset of graphs $\{ G_m \}_{m=1}^M$ from the labeled set $\mathcal{G}^L$ as the anchor graphs.
The corresponding constructed walk-induced hypergraphs and embedding-derived hypergraphs are $\{ H^w_m \}_{m=1}^{M}$ and $\{ H^e_m \}_{m=1}^M$, respectively.
We employ the hierarchical representation learning module to generate their representations as $\{ \mathbf{Z}^w_m \}_{m=1}^M$ and $\{ \mathbf{Z}^e_m \}_{m=1}^M$, respectively.
These anchor graph representations are relatively stable with label supervision compared to unlabeled graphs.
We maintain these representations in a queue initialized after 2 warm epochs and update the queue with labeled graphs from recent training iterations.
For a given unlabeled graph $G_u$ with walk-induced view representation $\mathbf{Z}_u^w$, we calculate the relational similarity between $\mathbf{Z}_u^w$ and all representations in $\{ \mathbf{Z}^w_m \}_{m=1}^M$ as the walk-induced view distribution $\mathbf{P}^u$ as follows:
\begin{equation}
\mathbf{P}^u_m = \frac{\mathrm{exp}\left\{\mathrm{sim}(\mathbf{Z}_u^w, \mathbf{Z}_m^w))  / \tau \right\}} {\sum_{m^\prime = 1}^M \mathrm{exp} \left\{ \mathrm{sim}(\mathbf{Z}^w_u, \mathbf{Z}_{m^\prime}^w) / \tau \right\}},
\end{equation}
where $\mathrm{sim}(\cdot)$ is the cosine similarity function, $\tau=0.5$ is a temperature parameter.
Similarly, we can calculate the embedding-derived view distribution $\mathbf{Q}^u$.
The consistency loss is designed to minimize the Kullback-Leibler (KL) Divergence between these two distributions as follows:
\begin{equation}\label{eq_lcon}
\mathcal{L}_\mathrm{con} = \frac{1}{2|\mathcal{G}^U|} \sum_{G^u \in \mathcal{G}^U} \left( D_{\mathrm{KL}}(\mathbf{P}^u \| \mathbf{Q}^u) + D_{\mathrm{KL}}(\mathbf{Q}^u \| \mathbf{P}^u) \right).
\end{equation}

\begin{algorithm}[t] 
\caption{Algorithm of HypSEE Framework}
\label{algo_main}
\begin{algorithmic}[1]
\STATE \textbf{Input:} Labeled graph set $\mathcal{G}^L$, unlabeled graph set $\mathcal{G}^U$;
\STATE \textbf{Output:} Trained graph classifier.
\STATE Initialize parameters in HypSEE.
\STATE Select $M$ labeled graphs from $\mathcal{G}^L$ as anchor graphs and store their representations in a memory bank.
\WHILE{not converge}
\STATE Sample minibatch $\mathcal{B}^L$ and $\mathcal{B}^U$.
\STATE Construct walk-induced hypergraphs and embedding-derived hypergraphs.
\STATE Calculate the representations of $\mathcal{B}^L$ and $\mathcal{B}^U$ using the hierarchical representation module.
\STATE Calculate $\mathcal{L}_\mathrm{hse}$ using Eq. (\ref{eq_lhse}).
\STATE Calculate $\mathcal{L}_\mathrm{con}$ using Eq. (\ref{eq_lcon}).
\STATE Calculate $\mathcal{L}_\mathrm{sup}$ using Eq. (\ref{eq_lsup}).
\STATE Update parameters by $\mathcal{L}_\mathrm{HypSEE}$ in Eq. (\ref{eq_overall_objective}).
\STATE Update graph representations in the memory bank.
\ENDWHILE

\end{algorithmic}
\end{algorithm}

\subsection{Optimization Framework}
We employ a two-layer MLP as the classifier $f_{\Xi}(\cdot)$ to predict graph labels according to the learned representations.
For any labeled graph $G_i \in \mathcal{G}^L$, the supervision from graph labels is brought into the model by the supervised loss $\mathcal{L}_{\mathrm{sup}}$ calculated as follows:
\begin{equation}\label{eq_lsup}
\mathcal{L}_{\mathrm{sup}} = -\frac{1}{|\mathcal{G}^L|} \sum_{i=1}^{|\mathcal{G}^L|} \mathbf{y}_i \log (p_i), \ \ \ \ p_i =  f_{\Xi}(\mathbf{Z}^w_i \mathbin{\oplus} \mathbf{Z}^e_i),
\end{equation}
where $\mathbin{\oplus}$ is the concatenation operator, $p_i$ is the predicted probability of $G_i$, and $\mathbf{y}_i$ is the corresponding ground truth label.
Finally, the overall objective function of HypSEE is formulated as follows:
\begin{equation}
\label{eq_overall_objective}
\mathcal{L}_{\mathrm{HypSEE}} = \mathcal{L}_\mathrm{sup} + \eta_1 \mathcal{L}_\mathrm{con} + \eta_2 \mathcal{L}_\mathrm{hse},
\end{equation}
where $\eta_1$ and $\eta_2$ are hyperparameters to balance loss components.

\begin{table*}[t]
\caption{Performance (ACC \%) of methods on 8 benchmark graph classification datasets, with standard deviations over 5 runs. Bold: the highest scores, and underline: the second-highest scores.}
\label{tab::results}
\centering
\tabcolsep=6.5pt
\begin{tabular}{l|cccccccc}
\toprule
{Methods} &  {PROTEINS} & {DD} & {IMDB-B} & {IMDB-M}   & {REDDIT-B} & {REDDIT-M-5K} & {COLLAB} & {MSRC21}  \\
\midrule 
WL         & 63.5 $\pm$ 1.6 & 57.3 $\pm$ 1.2 & 58.1 $\pm$ 1.2 & 33.3 $\pm$ 1.4 & 61.8 $\pm$ 1.3 & 37.0 $\pm$ 0.9 & 62.9 $\pm$ 0.7 & 30.7 $\pm$ 2.9  \\
Sub2Vec    & 52.7 $\pm$ 4.5 & 46.4 $\pm$ 3.2 & 44.9 $\pm$ 3.5 & 31.8 $\pm$ 2.7 & 63.5 $\pm$ 2.3 & 35.1 $\pm$ 1.5 & 60.8 $\pm$ 1.4 & 18.3 $\pm$ 6.2  \\
Graph2Vec  & 63.1 $\pm$ 1.8 & 53.7 $\pm$ 1.6 & 61.2 $\pm$ 2.6 & 38.1 $\pm$ 2.2 & 67.7 $\pm$ 2.3 & 38.1 $\pm$ 1.4 & 63.6 $\pm$ 0.9 & 28.4 $\pm$ 3.2 \\
\midrule
EntMin     & 62.7 $\pm$ 2.7 & 59.8 $\pm$ 1.3 & 67.1 $\pm$ 3.7 & 37.4 $\pm$ 1.2 & 66.9 $\pm$ 3.5 & 38.7 $\pm$ 2.8 & 63.8 $\pm$ 1.6 & 32.1 $\pm$ 3.3  \\
Mean-Teacher &64.3 $\pm$ 2.1 & 60.6 $\pm$ 1.8 & 66.4 $\pm$ 2.7 & 38.8 $\pm$ 3.6 & 68.7 $\pm$ 1.3 & 39.2 $\pm$ 2.1 & 63.6 $\pm$ 1.4 & 34.6 $\pm$ 2.6  \\
VAT        & 64.1 $\pm$ 1.2 & 59.9 $\pm$ 2.6 & 67.2 $\pm$ 2.9 & 39.6 $\pm$ 1.4 & 70.8 $\pm$ 4.1 & 38.9 $\pm$ 3.2 & 64.1 $\pm$ 1.1 & 33.8 $\pm$ 3.4  \\
\midrule
InfoGraph  & 68.2 $\pm$ 0.7 & 67.5 $\pm$ 1.4 & 71.8 $\pm$ 2.3 & 42.3 $\pm$ 1.8 & 75.2 $\pm$ 2.4 & 41.5 $\pm$ 1.7 & 65.7 $\pm$ 0.4 & 40.6 $\pm$ 2.4  \\
ASGN       & 67.7 $\pm$ 1.2 & 68.5 $\pm$ 0.6 & 70.6 $\pm$ 1.4 & 41.2 $\pm$ 1.4 & 73.1 $\pm$ 2.3 & 42.2 $\pm$ 0.8 & 65.3 $\pm$ 0.8 & 38.0 $\pm$ 2.5  \\
GraphCL    & 69.4 $\pm$ 0.8 & 68.7 $\pm$ 1.2 & 71.2 $\pm$ 2.5 & 43.7 $\pm$ 1.3 & 75.2 $\pm$ 1.7 & 42.3 $\pm$ 0.9 & 66.4 $\pm$ 0.6 & -  \\
JOAO       & 68.7 $\pm$ 0.9 & 67.9 $\pm$ 1.3 & 71.0 $\pm$ 1.9 & 42.6 $\pm$ 1.5& 74.8 $\pm$ 1.6 & 42.1 $\pm$ 1.2 & 65.8 $\pm$ 0.4 & 35.4 $\pm$ 2.5  \\
DualGraph  & 70.1 $\pm$ 1.2 & 69.8 $\pm$ 0.8 & 72.1 $\pm$ 0.7 & \underline{44.8 $\pm$ 0.4} & 75.4 $\pm$ 1.4 & 42.9 $\pm$ 1.4 & 67.2 $\pm$ 0.6 & 37.3 $\pm$ 2.3 \\
KGNN       & 70.9 $\pm$ 0.5 & 70.5 $\pm$ 0.6 & 72.5 $\pm$ 1.6 & 43.3 $\pm$ 0.7 & 76.0 $\pm$ 0.9 & 44.8 $\pm$ 0.6 & 67.4 $\pm$ 0.5 & -  \\
TGNN       & 71.0 $\pm$ 0.7 & 70.8 $\pm$ 0.9 & 72.8 $\pm$ 1.7 & 42.9 $\pm$ 0.8 & 76.3 $\pm$ 1.3 & 43.8 $\pm$ 1.0 & 67.7 $\pm$ 0.4 & -  \\
GraphSpa   & 71.2 $\pm$ 0.7 & 71.4 $\pm$ 0.8 & 72.3 $\pm$ 1.1 & 44.5 $\pm$ 0.6 & 76.5 $\pm$ 0.4 & 44.0 $\pm$ 0.6 & - & -  \\
HEAL  &  \underline{73.4 $\pm$ 0.8}  &  \underline{72.1 $\pm$ 0.9} & \textbf{73.5 $\pm$ 1.5} & 44.3 $\pm$ 0.6  & -    &  45.9 $\pm$ 1.0  &  68.3 $\pm$ 0.5 & -  \\
DisenSemi & 67.5 $\pm$ 2.7 & 68.3 $\pm$ 1.5 & 70.2 $\pm$ 3.9 & 41.7 $\pm$ 7.8 & \underline{87.8 $\pm$ 1.3} & \underline{51.8 $\pm$ 2.2} & \underline{71.5 $\pm$ 1.5} & \underline{61.1 $\pm$ 3.0} \\
\midrule
HypSEE & \textbf{73.5 $\pm$ 3.8} & \textbf{74.0 $\pm$ 1.4} & \underline{73.0 $\pm$ 2.2}& \textbf{45.5 $\pm$ 3.3} & \textbf{88.1 $\pm$ 1.9} & \textbf{52.4 $\pm$ 0.5} & \textbf{72.0 $\pm$ 1.7} & \textbf{69.2 $\pm$ 5.3} \\
\bottomrule
\end{tabular}

\end{table*}

\noindent\textbf{Algorithm.}
We summarize the main procedures in HypSEE in Algorithm \ref{algo_main}.
For hypergraph construction, building the random walk-induced and embedding-derived hypergraphs requires $O(K^w l_\epsilon)$ and $O(|\mathcal{E}| + K^e d |\mathcal{V}|)$ time, respectively.
These walk-induced hypergraphs are precomputed offline to eliminate the computational overhead of sequential random walks per epoch.
In the hierarchical representation learning module, hypergraph convolution, hypergraph pooling, and the calculation of the HSE loss $\mathcal{L}_\mathrm{hse}$ are executed level-by-level.
Specifically, at each depth $h$, these three operations incur complexities of $O(N_h(K^w+K^e)d)$, $O(N_h d N_{h-1})$, and $O((N_h)^2(K^w+K^e))$, assuming a small partitioning tree depth $K$.
Summing these across all levels yields a cumulative complexity of $O( (K^w+K^e)|\mathcal{V}|(d+|\mathcal{V}|))$ for this module.
Furthermore, computing the consistency loss $\mathcal{L}_{\mathrm{con}}$ takes $O(Md)$ time.
Overall, the total time complexity of HypSEE evaluates to $O(K^w l_\epsilon + |\mathcal{E}| + (K^w+K^e)|\mathcal{V}|(d+|\mathcal{V}|) + Md)$, which is asymptotically dominated by $O((K^w+K^e)|\mathcal{V}|^2)$.

\begin{table}[t]
\centering
\caption{Details of dataset statistics.}
\label{tab::datasets}
\begin{tabular}{lccccc}
\toprule
Datasets & Graphs & Nodes & Edges & Fea. & Classes \\
\midrule
PROTEINS    & 1113 & 39.06  & 72.82   & 3  & 2  \\
DD          & 1178 & 284.32 & 715.66  & 89 & 2  \\
IMDB-B      & 1000 & 19.77  & 96.53   & 0  & 2  \\
IMDB-M      & 1500 & 13.00  & 65.94   & 0  & 3  \\
REDDIT-B    & 2000 & 429.63 & 497.75  & 0  & 2  \\
REDDIT-M-5K & 4999 & 508.52 & 594.87  & 0  & 5  \\
COLLAB      & 5000 & 74.49  & 2457.78 & 0  & 3  \\
MSRC21      & 563  & 77.52  & 198.32  & 24 & 20 \\
\bottomrule
\end{tabular}
\end{table}

\section{Experiments}
\subsection{Experimental Setup}
\noindent\textbf{Datasets.}
We conduct experiments on 8 well-known benchmark graph datasets from the TUDataset \cite{morris2020tudataset}.
They include 2 bioinformatics datasets (PROTEINS \cite{borgwardt2005protein} and DD \cite{dobson2003distinguishing}), 5 social network datasets (IMDB-BINARY, IMDB-MULTI, REDDIT-B, REDDIT-M-5K, and COLLAB \cite{yanardag2015deep}), and 1 computer vision dataset (MSRC21 \cite{neumann2016propagation}).
Detailed statistics of these datasets are summarized in Table \ref{tab::datasets}.
To achieve fair comparisons, we split the dataset into labeled and unlabeled training sets, a validation set, and a test set, with a ratio of 2:5:1:2, following DualGraph \cite{luo2022dualgraph}.
We use only half of the labeled training graphs, i.e., 10\% of the dataset, for training.

\noindent\textbf{Baselines.}
We compare HypSEE against baselines from three categories: classic graph representation methods, general semi-supervised learning methods, and SSGC methods.
For classic graph representation methods, we consider WL \cite{shervashidze2011weisfeiler}, Graph2Vec \cite{narayanan2017graph2vec}, and Sub2Vec \cite{adhikari2018sub2vec}.
For general semi-supervised learning methods, we consider EntMin \cite{grandvalet2004semi}, Mean-Teacher \cite{tarvainen2017mean}, and VAT \cite{miyato2018virtual}.
For SSGC methods, we consider InfoGraph \cite{suninfograph}, GraphCL \cite{you2020graph}, ASGN \cite{hao2020asgn}, JOAO \cite{you2021graph}, DualGraph \cite{luo2022dualgraph}, KGNN \cite{ju2022kgnn}, TGNN \cite{ju2022tgnn}, GraphSpa \cite{ju2024focus}, HEAL \cite{ju2024hypergraph}, and DisenSemi \cite{wang2025disensemi}.

\noindent\textbf{Implementation Details.} 
HypSEE is developed in PyTorch and optimized end-to-end using the Adam optimizer.
Specifically, the training process is conducted with a learning rate of $0.001$, a weight decay of $1\times 10^{-4}$, and a batch size of $80$ for $100$ epochs. 
For the hypergraph construction, both views share $K^w = K^e = 64$ hyperedges, with lengths of $\{4, 6\}$ for the walk-induced view and a dimension of $d^e=128$ for the embedding-derived view. 
The hidden dimension of the hypergraph convolution is fixed at $d=64$. 
Furthermore, the hierarchical module is configured with a depth of $K=2$, a node reduction rate of $0.3$, and $M=64$ anchor graphs. 
Finally, the trade-off hyperparameters in the overall objective are controlled by $\eta_1=1$ and $\eta_2=0.001$.
The source code is available at GitHub\footnote{https://github.com/SELGroup/HypSEE}.

\subsection{Experimental Results}
Table \ref{tab::results} reports the classification performance of all competing methods across the 8 datasets. Based on these results, we draw the following key observations:
\textbf{(1) Superiority of HypSEE.} HypSEE consistently achieves state-of-the-art performance, securing the highest accuracy on 7 datasets and the runner-up position on IMDB-BINARY. 
We attribute this remarkable efficacy to three core mechanisms: 
(i) both the walk-induced and embedding-derived hypergraph views successfully capture higher-order graph topologies, inherently mitigating the over-smoothing issue prevalent in traditional GNNs; 
(ii) the HSE-guided hierarchical module explicitly models the graph's hierarchical structures; 
and (iii) the anchor graph-based consistency learning aligns the two distinct views, allowing them to mutually enhance each other. 
Notably, HypSEE exhibits particularly strong performance on larger-scale graphs, implying its capability to exploit pronounced hierarchical properties. 
Conversely, on IMDB-BINARY—which consists of extremely small graphs (averaging 20 vertices), HypSEE is slightly outperformed by HEAL, likely because the hierarchical structural cues in such small topologies are less distinct.
\textbf{(2) Baseline Analysis.} Compared to classic graph representation and generic semi-supervised learning baselines, SSGC-tailored methods demonstrate significantly better performance, corroborating the necessity of graph-specific designs. 
Among the baselines, HEAL achieves the second-best overall performance, further validating the utility of leveraging hypergraphs for higher-order relationship modeling. 
Meanwhile, although DisenSemi shows promise on several large-graph datasets, its performance fluctuates considerably on others, indicating limited cross-dataset robustness. 

\noindent\textbf{Large Scale Datasets.}
We further conduct experiments on two OGB molecular datasets: ogbg-molhiv (HIV; 41,127 graphs, HIV inhibition prediction) and ogbg-molmuv (MUV; 93,087 graphs, 17 virtual-screening tasks).
As reported in Table \ref{tab:hiv_muv}, HypSEE achieves the highest test ROC-AUC in three of four settings. 
At 10\% labels, it exceeds the strongest baseline, DisenSemi, by 2.0 and 4.9 percentage points on HIV and MUV, respectively. At 1\%, it leads on HIV and performs comparably to DisenSemi on MUV (60.6\% vs. 60.7\%). 
These results support its effectiveness on larger graph collections under limited supervision.
Collectively, HypSEE establishes a robust advantage over all competitors by synergistically modeling both high-order and hierarchical graph structures.

\begin{figure}[t]
\centering
\begin{subfigure}{0.49\linewidth}
    \centering
    \captionsetup{font=scriptsize}
    \caption*{PROTEINS}
    \includegraphics[width=\linewidth,trim={0 0 0 0}, clip]{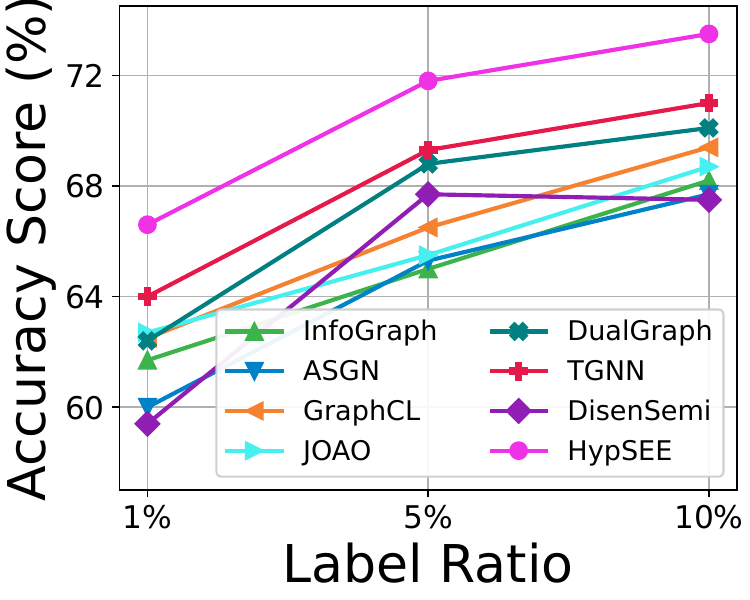}
\end{subfigure}
\begin{subfigure}{0.49\linewidth}
    \centering
    \captionsetup{font=scriptsize}
    \caption*{DD}
    \includegraphics[width=\linewidth,trim={0 0 0cm 0cm}, clip]{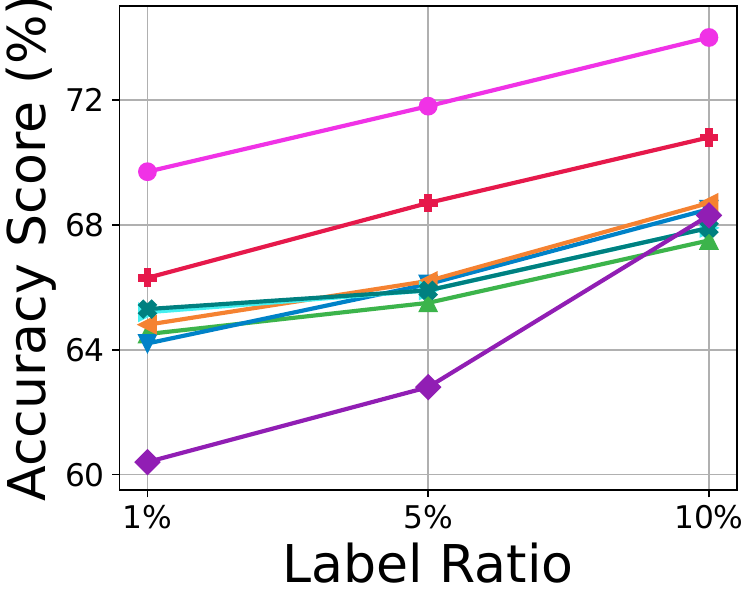}
\end{subfigure}

\caption{Performance results with different labeling ratios.
}
\Description{Performance results with different labeling ratios.}
\label{fig_label_ratio}
\end{figure}

\noindent\textbf{Influence of Labeling Ratio.}
To evaluate model robustness under varying degrees of supervision, we conduct experiments on the PROTEINS and DD datasets with different labeling ratios (i.e., the proportion of labeled training graphs). 
As depicted in Fig. \ref{fig_label_ratio}, while the accuracy of all methods naturally improves with increasing labeling ratio, HypSEE consistently maintains its superiority across all settings. 
This sustained performance, even in low-label regimes, highlights the effectiveness of our dual-view higher-order and hierarchical structure modeling. 
Overall, HypSEE demonstrates exceptional robustness and adaptability to label scarcity, confirming its practical utility for SSGC.

\begin{table}[t]
    \centering
    \setlength{\tabcolsep}{14pt}
    \begin{tabular}{@{}llcc@{}}
        \toprule
        Label rate & Methods & HIV & MUV \\
        \midrule
        \multirow{5}{*}{1\%}
        & InfoGraph & $51.3 \pm 3.8$ & $50.1 \pm 1.6$ \\
        & GraphCL   & $45.1 \pm 5.7$ & $50.8 \pm 1.3$ \\
        & TGNN      & $53.3 \pm 4.1$ & $52.8 \pm 1.5$ \\
        & DisenSemi & $60.2 \pm 3.8$ & $\mathbf{60.7} \pm \mathbf{1.1}$ \\
        \cmidrule{2-4}
        & HypSEE    & $\mathbf{60.8} \pm \mathbf{3.1}$ & $60.6 \pm 3.3$ \\
        \midrule
        \multirow{5}{*}{10\%}
        & InfoGraph & $63.7 \pm 1.2$ & $51.3 \pm 0.8$ \\
        & GraphCL   & $63.2 \pm 0.5$ & $51.1 \pm 0.7$ \\
        & TGNN      & $64.1 \pm 0.8$ & $53.5 \pm 0.7$ \\
        & DisenSemi & $70.4 \pm 3.2$ & $62.4 \pm 1.7$ \\
        \cmidrule{2-4}
        & HypSEE    & $\mathbf{72.4} \pm \mathbf{0.8}$
                    & $\mathbf{67.3} \pm \mathbf{3.1}$ \\
        \bottomrule
    \end{tabular}
    \caption{Results (test ROC-AUC \%) on large-scale OGB datasets HIV and MUV at 1\% and 10\% label rates.}
    \label{tab:hiv_muv}
\end{table}

\subsection{Ablation Study}

\begin{table}[t]
\centering
\caption{Experimental results (ACC \%) of ablation study.}
\label{tab::ablation}
\newcommand{\std}[1]{{$\pm$#1}}

\begin{tabular}{lcccc}
\toprule
Methods & PROTEINS & IMDB-M & COLLAB & MSRC21  \\
\midrule
Walk-Single  & 71.8\std{3.8} & 45.0\std{3.4} & 71.0\std{2.2} & 67.1\std{6.7} \\
Embed-Single  & 73.1\std{3.0} & 42.6\std{3.5}& 71.7\std{1.2} & 68.7\std{6.0}  \\
w/o Pooling & 71.8\std{4.3} & 44.1\std{2.9} & 71.8\std{1.5} & 68.8\std{4.9} \\
w/o $\mathcal{L}_\mathrm{hse}$ & 72.2\std{4.7} & 45.3\std{2.2} & 71.4\std{1.5} & 69.0\std{4.3} \\
w/o $\mathcal{L}_\mathrm{con}$ & 72.6\std{4.1} & 42.5\std{4.3} & 70.6\std{2.5} & 68.3\std{5.5} \\
\midrule
HypSEE & \textbf{73.5}\textbf{\std{3.8}} & \textbf{45.5\std{3.3}} & \textbf{72.0}\textbf{\std{1.7}} & \textbf{69.2}\textbf{\std{5.3}} \\
\bottomrule
\end{tabular}

\end{table}

To verify the effectiveness of the core components in HypSEE, we conduct an ablation study with results reported in Table \ref{tab::ablation}. 
Specifically, we design the following five variants: (1) \textbf{Walk-Single} and (2) \textbf{Embed-Single}, which rely solely on the walk-induced or embedding-derived hypergraph view, respectively; (3) \textbf{HypSEE w/o Pooling}, which removes the hierarchical pooling module; and two variants omitting specific optimization objectives, namely (4) \textbf{HypSEE w/o $\mathcal{L}_\mathrm{con}$} and (5) \textbf{HypSEE w/o $\mathcal{L}_\mathrm{hse}$}.
As observed in Table \ref{tab::ablation}, relying on any single hypergraph view yields sub-optimal performance. 
While simply combining the two views (i.e., w/o $\mathcal{L}_\mathrm{con}$) provides more stable representations, the performance remains limited without the explicit cross-view alignment provided by the consistency loss. 
Furthermore, the noticeable performance drops in variants lacking the hierarchical pooling or its guiding loss ($\mathcal{L}_\mathrm{hse}$) confirm that explicitly modeling HSE-guided hierarchical structures is crucial for learning comprehensive graph representations.

\begin{figure}[t]
\centering
\begin{subfigure}{0.49\linewidth}
    \centering
    \includegraphics[width=\linewidth,trim={0 0 0 0}, clip]{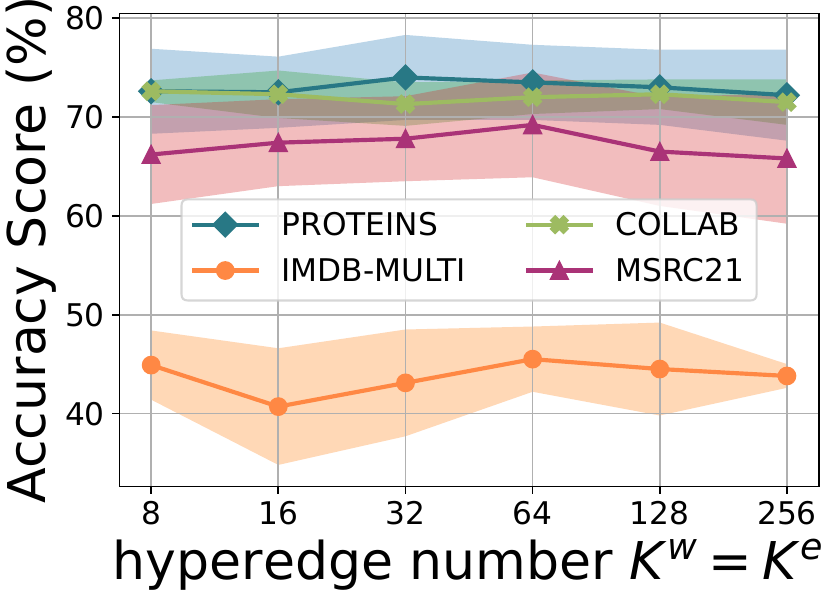}
\end{subfigure}
\begin{subfigure}{0.49\linewidth}
    \centering
    \includegraphics[width=\linewidth,trim={0 0 0cm 0cm}, clip]{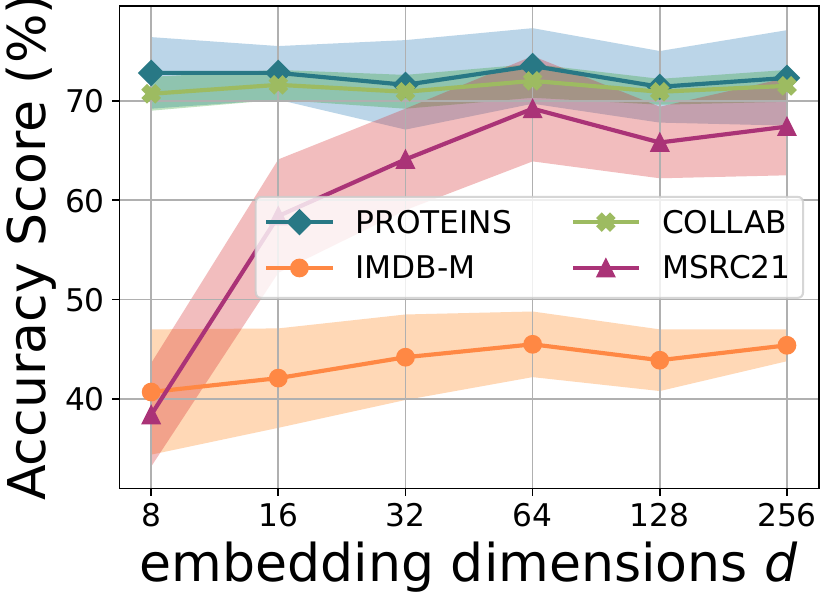}
\end{subfigure}

\caption{Parameter sensitivity analysis on hyperedge number $K^w=K^e$ and embedding dimension $d$.
}
\Description{Parameter sensitivity analysis plots.}
\label{fig_sensitivity}
\end{figure}

\begin{figure}[t]
\centering
\begin{subfigure}{0.49\linewidth}
    \centering
    \includegraphics[width=\linewidth,trim={0 0 0 0}, clip]{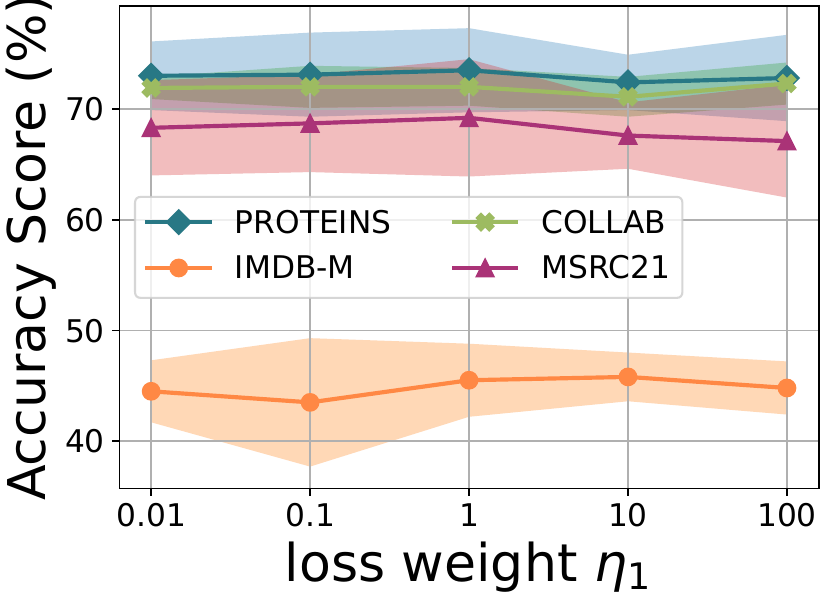}
\end{subfigure}
\begin{subfigure}{0.49\linewidth}
    \centering
    \includegraphics[width=\linewidth,trim={0 0 0cm 0cm}, clip]{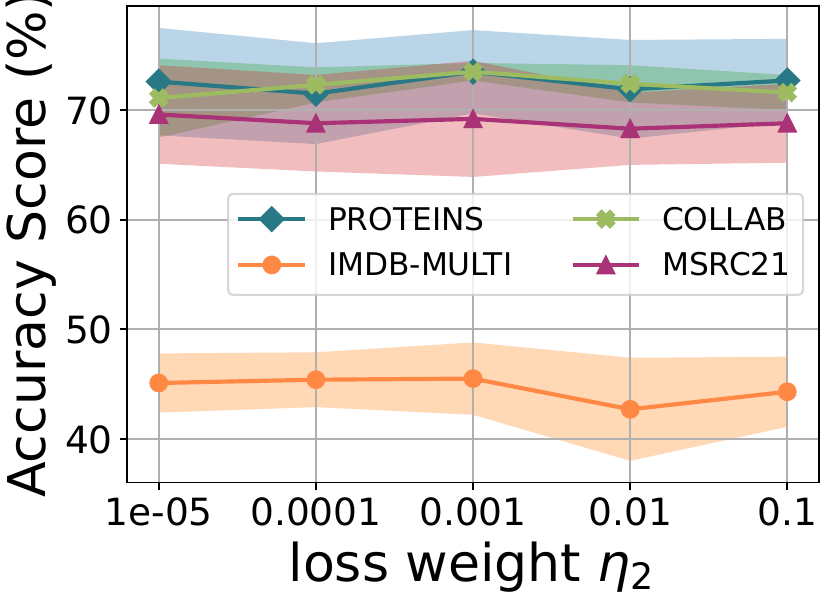}
\end{subfigure}

\caption{Parameter sensitivity on loss weights $\eta_1$ and $\eta_2$.
}
\Description{Parameter sensitivity analysis on loss weights.}
\label{fig_sensitivity_loss}
\end{figure}

\begin{figure}[t]
\centering
\includegraphics[width=1.0\linewidth]{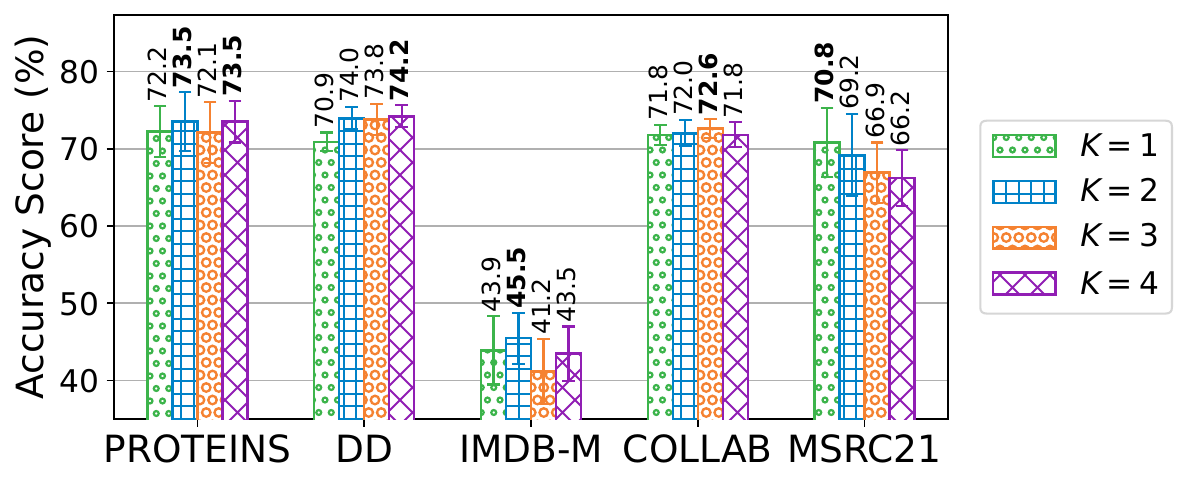} 
\caption{Parameter sensitivity analysis on tree height $K$.}
\Description{Tree heights.}
\label{fig_height}
\end{figure}

\subsection{Hyperparameter Sensitivity}

\noindent\textbf{Hyperedge Numbers \& Embedding Dimensions.}
We investigate the impact of hyperedge numbers ($K^w=K^e$) and embedding dimensions ($d$) across 4 datasets (Fig. \ref{fig_sensitivity}). 
Generally, HypSEE maintains stable performance across varying hyperedge numbers. 
It achieves peak accuracy around $K^w=K^e=64$, confirming that $64$ hyperedges are sufficient for high-order structure extraction. 
The minor instability on MSRC21 suggests its topological patterns are harder to capture.
To balance optimal performance and computational efficiency, $K^w=K^e=64$ is adopted as the default. 
Regarding the embedding dimension $d$, performance remains stable on PROTEINS and COLLAB, but improves with larger $d$ on IMDB-M and MSRC21, indicating a need for greater representational capacity. 
Considering the rising computational costs associated with larger dimensions, we select $d=64$ as the default trade-off.

\noindent\textbf{Loss weights.}
To balance the components of the overall objective in Eq. (\ref{eq_overall_objective}), we introduce two hyperparameters: $\eta_1$ and $\eta_2$.
As depicted in Figure \ref{fig_sensitivity_loss}, HypSEE demonstrates general robustness to these weights, with slightly different performance trends.
As the consistency learning loss weight $\eta_1$ increases, the model's performance initially improves but subsequently declines, reaching an optimal balance at $\eta_1=1$. 
Notably, assigning an excessively large value to $\eta_1$ leads to a marked performance drop on the MSRC dataset, implying that overpowering the primary objective is detrimental.
Similarly, for the HSE loss weight $\eta_2$, empirical results show a performance peak at $\eta_2=0.001$ on PROTEINS and COLLAB. 
Consequently, we set the default values to $\eta_1=1$ and $\eta_2=0.001$, as this configuration consistently yields high performance across datasets.

\noindent\textbf{Partitioning tree height.}
We investigate the impact of partitioning tree height $K$, which determines the number of hypergraph convolution layers and influences the HSE loss $\mathcal{L}_{\mathrm{hse}}$.
While the optimal tree height varies across datasets, HypSEE consistently maintains highly competitive performance at $K=2$.
Notably, the performance degrades significantly at $K=1$ on several datasets, particularly on DD.
This highlights the necessity of hierarchical pooling, as a flat structure ($K=1$) fails to provide meaningful hierarchical guidance for the model.

\noindent\textbf{Hypergraph construction.}
To generate the walk-induced view, we sample $K^w$ starting vertices with replacement for each graph and perform random walks of length $l_w$.
Table \ref{tab_hyperedge_length} analyzes the impact of the hyperedge length $l_w$ on the overall performance of HypSEE.
Experimental results demonstrate that configuring the sampled lengths to $\{4,6\}$ or $\{5,5\}$ consistently yields optimal performance.

\begin{table}[t]
\caption{Parameter sensitivity on hyperedge length (ACC \%).} 
\label{tab_hyperedge_length}
\centering
\begin{tabular}{cccccc}
\toprule
 $l_w$ & PROTEINS & DD & COLLAB & MSRC21 \\
 \midrule
  $\{5,5\}$ & 73.4$\pm$3.4 & 75.5$\pm$2.4 & 71.6$\pm$1.6 & 69.7$\pm$5.0 \\
 \cellcolor{gray!25}$\{4,6\}$ & \cellcolor{gray!25}73.5$\pm$3.8 & \cellcolor{gray!25}74.0$\pm$1.4 & \cellcolor{gray!25}72.0$\pm$1.7 & \cellcolor{gray!25}69.2$\pm$5.3 \\
 $\{3,7\}$ & 71.2$\pm$5.1 & 72.6$\pm$3.5 & 70.9$\pm$1.5 & 69.0$\pm$6.1 \\
 $\{2,8\}$ & 72.6$\pm$4.1 & 73.6$\pm$1.8 & 71.3$\pm$1.7 & 68.7$\pm$5.8 \\

\bottomrule
\end{tabular}
\end{table}

\begin{table}[t]
\caption{Parameter sensitivity on anchor graphs (ACC \%).} 
\label{tab_anchor}
\centering
\begin{tabular}{cccccc}
\toprule
 && PROTEINS & DD & COLLAB & MSRC21 \\
 \midrule
 \multirow{5}{*}{\rotatebox{90}{\makecell[c]{\#anchor}}} 
 & 16 & 72.9$\pm$3.7 & 73.6$\pm$2.0 & 72.1$\pm$2.0 & 68.8$\pm$3.6 \\
 & 32 & 72.3$\pm$4.2 & 73.6$\pm$1.1 & 72.3$\pm$1.7 & 67.6$\pm$3.9 \\
 & \cellcolor{gray!25}64 & \cellcolor{gray!25}73.5$\pm$3.8 & \cellcolor{gray!25}74.0$\pm$1.4 & \cellcolor{gray!25}72.0$\pm$1.7 & \cellcolor{gray!25}69.2$\pm$5.3 \\
 & 128 & 73.5$\pm$4.3 & 73.2$\pm$2.7 & 71.7$\pm$0.7 & 65.7$\pm$4.5 \\
 & 256 & 73.3$\pm$4.8 & 73.6$\pm$2.1 & 71.2$\pm$1.5 & 67.4$\pm$5.3 \\
\midrule

\multirow{5}{*}{\rotatebox{90}{\makecell[c]{\#warm epoch}}} 
& 0 & 71.7$\pm$4.9 & 73.5$\pm$2.1 & 71.5$\pm$1.9 & 69.0$\pm$5.8\\
& \cellcolor{gray!25}2 & \cellcolor{gray!25}73.5$\pm$3.8 & \cellcolor{gray!25}74.0$\pm$1.4 & \cellcolor{gray!25}72.0$\pm$1.7 & \cellcolor{gray!25}69.2$\pm$5.3 \\
& 4 & 71.7$\pm$4.0 & 74.3$\pm$1.0 & 72.1$\pm$1.0 & 67.4$\pm$5.8\\
& 6 & 72.8$\pm$3.4 & 74.1$\pm$1.2 & 71.8$\pm$1.3 & 67.8$\pm$4.8\\
& 8 & 71.7$\pm$3.8 & 75.2$\pm$1.9 & 72.6$\pm$2.0 & 68.1$\pm$5.9\\

\bottomrule
\end{tabular}
\end{table}

\noindent\textbf{Anchor Graphs.}
To avoid suboptimal initial representations, anchor graphs are updated during training but are only utilized after a warm-up phase.
Table \ref{tab_anchor} evaluates the impact of the anchor count and warm-up epochs.
With performance remaining largely stable across datasets, setting 64 anchors yields optimal results and is chosen as the default.
Furthermore, since HypSEE peaks at 2 or 4 warm-up epochs, we adopt 2 epochs as our default configuration.

\begin{figure}[t]
\centering
\begin{subfigure}{.15\textwidth}
    \centering
    \includegraphics[width=1.0\linewidth,trim={1.1cm 1.1cm 1.1cm 1.1cm}, clip]{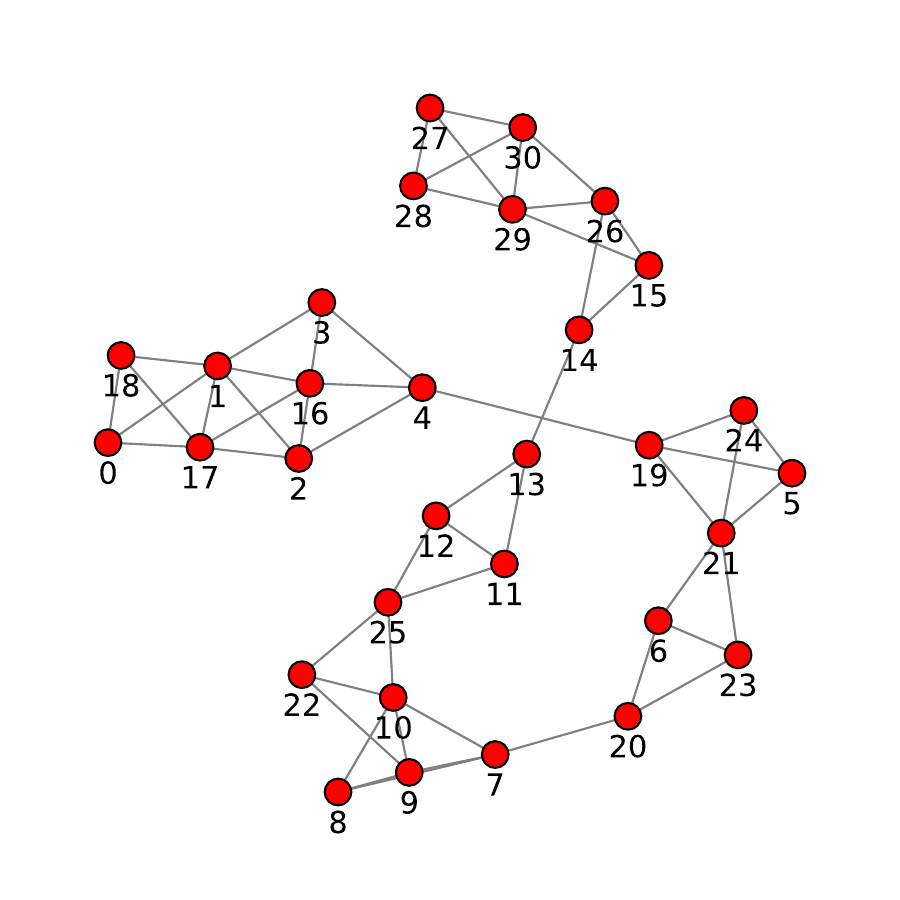}  
    \caption{Original graph}
    \label{fig_vis_graph}
\end{subfigure}
\begin{subfigure}{.15\textwidth}
    \centering
    \includegraphics[width=1.0\linewidth,trim={1.1cm 1.1cm 1.1cm 1.1cm}, clip]{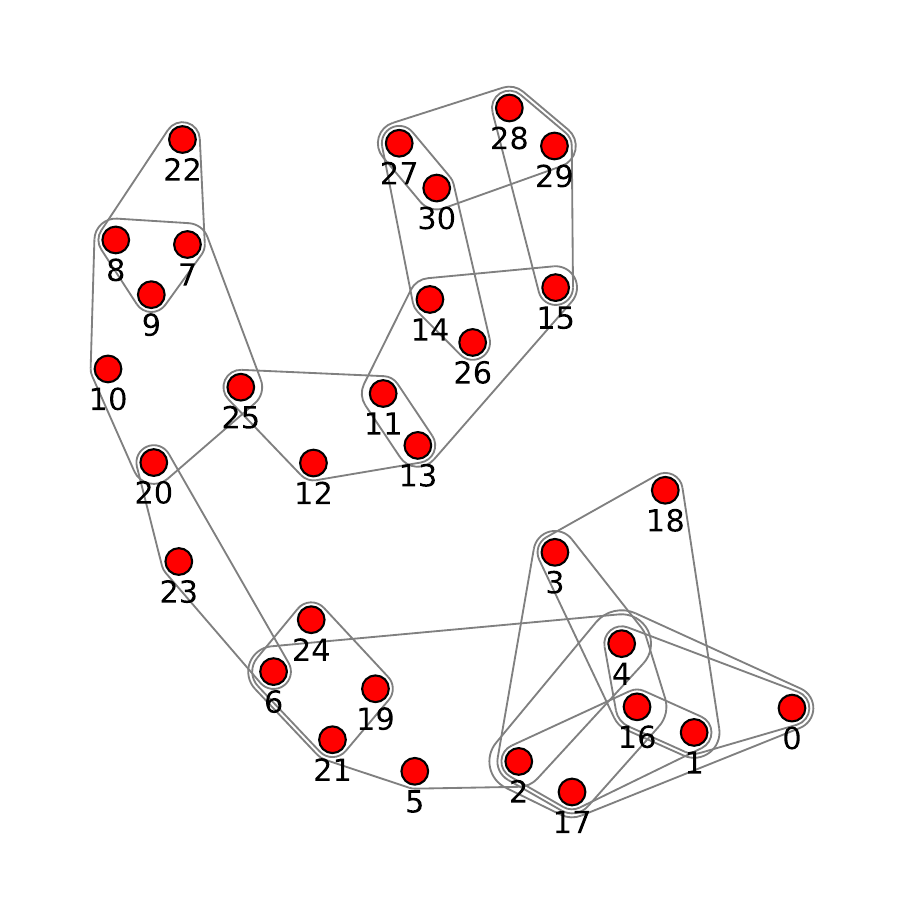}  
    \caption{Hypergraph 1}
    \label{fig_vis_hypergraph1}
\end{subfigure}
\begin{subfigure}{.15\textwidth}
    \centering
    \includegraphics[width=1.0\linewidth,trim={1.1cm 1.1cm 1.1cm 1.1cm}, clip]{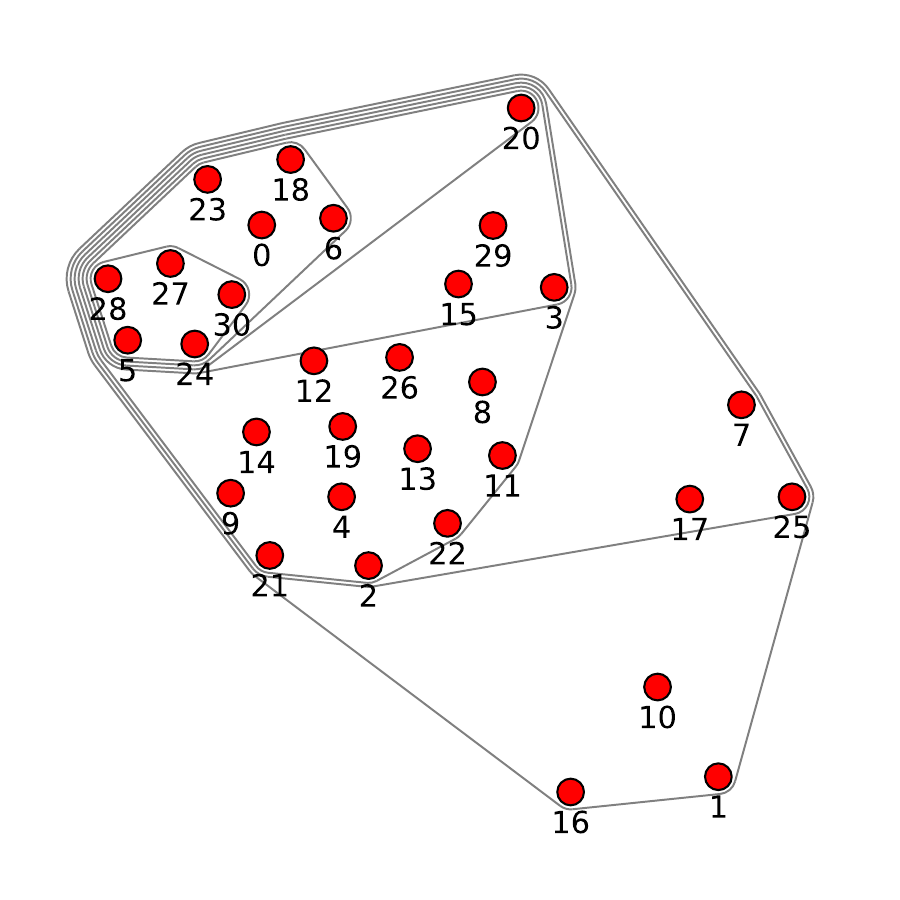}  
    \caption{Hypergraph 2}
    \label{fig_vis_hypergraph2}
\end{subfigure}
\caption{Visualization of (a) graph, (b) walk-induced hypergraph, and (c) embedding-derived hypergraph.}
\Description{Visualization of graphs and hypergraphs.}
\label{fig_vis}
\end{figure}

\subsection{Computational efficiency.}
To evaluate the computational efficiency of HypSEE, we compare it with DisenSemi on 8 datasets using an NVIDIA GeForce RTX 4090 GPU.
Table~\ref{tab:computational_efficiency} reports the training time per epoch and peak GPU memory usage.
HypSEE consistently achieves lower training time and memory consumption than DisenSemi across all evaluated datasets.
On REDDIT-MULTI-5K, HypSEE reduces the training time per epoch from 87.66 to 10.90 seconds and peak GPU memory usage from 22.64 to 12.76 GiB.
These results demonstrate that HypSEE offers computational advantages in both training speed and memory efficiency.

\begin{table}[t]
    \centering
    \begin{tabular}{@{}lcccc@{}}
        \toprule
        Dataset
        & \multicolumn{2}{c}{Time per epoch (s)}
        & \multicolumn{2}{c}{GPU memory usage (GiB)} \\
        \cmidrule(lr){2-3}
        \cmidrule(l){4-5}
        & HypSEE & DisenSemi & HypSEE & DisenSemi \\
        \midrule
        IMDB-B
        & \textbf{0.91}  & 1.66
        & \textbf{0.69}  & 4.74  \\
        IMDB-M
        & \textbf{1.31}  & 1.72
        & \textbf{0.68}  & 2.54  \\
        MSRC21
        & \textbf{0.65}  & 2.31
        & \textbf{1.55}  & 4.48  \\
        PROTEINS
        & \textbf{1.08}  & 2.56
        & \textbf{1.01}  & 3.54  \\
        DD
        & \textbf{1.31}  & 12.81
        & \textbf{4.95}  & 19.77 \\
        COLLAB
        & \textbf{4.42}  & 19.02
        & \textbf{5.89}  & 19.05 \\
        REDDIT-B
        & \textbf{10.39} & 29.73
        & \textbf{10.48} & 23.10 \\
        REDDIT-M
        & \textbf{10.90} & 87.66
        & \textbf{12.76} & 22.64 \\
        \bottomrule
    \end{tabular}
    \caption{Computational efficiency of HypSEE and DisenSemi.}
    \label{tab:computational_efficiency}
\end{table}

\subsection{Visualization Analysis.}
Fig. \ref{fig_vis} visualizes the walk-induced and embedding-derived hypergraphs of a sample from the PROTEINS dataset. Since the embedding-derived hypergraph is dense (with weights in $[0,1]$), we zero out weights below the median to enhance clarity. 
In Figs. \ref{fig_vis_hypergraph1}-\ref{fig_vis_hypergraph2}, hyperedges are depicted as vertices enclosed by boundaries. 
Unlike the original graph (Fig. \ref{fig_vis_graph}), which is restricted to pairwise connections, our hypergraphs capture higher-order relations among multiple nodes. 
Specifically, the walk-induced hypergraph (Fig. \ref{fig_vis_hypergraph1}) reveals local higher-order dependencies across varying walk lengths. 
In contrast, the embedding-derived hypergraph (Fig. \ref{fig_vis_hypergraph2}) excels at capturing long-range interactions, connecting nodes that are distant in the original topology.

\section{Related Works}
\noindent\textbf{Semi-Supervised Graph Classification.} 
Current SSGC methods are predominantly driven by two mechanisms. 
The first is pseudo-labeling (e.g., DualGraph \cite{luo2022dualgraph}, KGNN \cite{ju2022kgnn}, UGNN \cite{luo2023towards}), predicated on EM-style or prototype-aware training.
The second is consistency regularization, where strategies diverge into augmentation-based methods (GraphGL \cite{you2020graph}, JOAO \cite{you2021graph}, GLA \cite{yue2022label}) and view-alignment approaches (InfoGraph \cite{suninfograph}, TGNN \cite{ju2022tgnn}, GHNN \cite{ju2022ghnn}) that minimize divergence across structural views.
Closely related is HEAL \cite{ju2024hypergraph}, which enforces consistency between higher-order hypergraphs and induced line graphs.
However, existing SSGC frameworks fundamentally overlook hierarchical graph organizations.

\noindent\textbf{Hypergraph Learning.}
Hypergraphs capture higher-order, multi-way relationships by allowing hyperedges to connect multiple vertices \cite{kim2024survey}.
Fundamental tasks like hypergraph partitioning have been extensively addressed using spectral clustering \cite{zhou2006learning}, PageRank \cite{takai2020hypergraph}, and modularity \cite{chodrow2021generative}.
Recently, hypergraph neural networks (HGNNs) have generalized GNNs to model higher-order structures effectively \cite{feng2019hypergraph,meng2026DHMNN,ju2024hypergraph,zhang2025enhanced}.
In HypSEE, we enable HGNNs to learn hierarchical representations through an HSE-guided pooling layer.

\noindent\textbf{Hierarchy-Aware Graph Learning.}
Graphs naturally contain hierarchical structures that reveal crucial vertex influence patterns \cite{wu2024hierarchy}.
While traditional methods rely on hierarchical clustering \cite{dasgupta2016cost} or community detection \cite{lancichinetti2009detecting}, recent GNNs address this using hyperbolic spaces \cite{chami2019hyperbolic} or ranking-based node-level representations like HAGNN \cite{wu2024hierarchy}.
However, these node-centric methods incur high computational costs and have unverified graph-level capabilities.
Conversely, hierarchical graph pooling \cite{ying2018hierarchical,bianchi2020spectral,yang2024sebot} captures hierarchies for graph classification but is limited to supervised settings.

\section{Conclusion}
In this paper, we present HypSEE, a novel hierarchy-aware approach for SSGC built upon HSE. 
We first derive two distinct hypergraph views: walk-induced hypergraphs and embedding-derived hypergraphs, to capture complex, higher-order relationships within graphs. 
To ensure a hierarchical representation in these hypergraphs, we propose HSE to guide hypergraph coarsening, thereby effectively capturing the underlying hierarchical structures. 
Finally, we refine representations from both views using a graph anchor-based consistency learning strategy, enabling robust learning from unlabeled graphs. 
Extensive experiments conducted across 8 datasets showcase HypSEE's SOTA performance in SSGC.

\begin{acks}
This work was supported by the Central Government-Guided Local Science and Technology Development Fund Project of the Hebei Provincial Department of Science and Technology (Grant No. $\ \ $ 254Z9902G), the Key Project of the Joint Fund of the National Natural Science Foundation of China (NSFC) (Grant No. U25B2029), the Beijing Natural Science Foundation–Xiaomi Innovation Joint Fund (Cultivation Project) (Grant No. L253021), the Major Science and Technology Special Program of the Yunnan Provincial Department of Science and Technology (Grant Nos. 202502AD080012 and 202502AD080006), and the CCF-DiDi Collaborative Project (Grant No. 202614).
\end{acks}

\clearpage

\bibliographystyle{ACM-Reference-Format}
\bibliography{references}

@inproceedings{ying2018hierarchical,
  title={Hierarchical graph representation learning with differentiable pooling},
  author={Ying, Rex and You, Jiaxuan and Morris, Christopher and Ren, Xiang and Hamilton, William L and Leskovec, Jure},
  booktitle={Proceedings of NeurIPS},
  pages={4805--4815},
  year={2018}
}

@inproceedings{luo2022dualgraph,
  title={Dualgraph: Improving semi-supervised graph classification via dual contrastive learning},
  author={Luo, Xiao and Ju, Wei and Qu, Meng and Chen, Chong and Deng, Minghua and Hua, Xian-Sheng and Zhang, Ming},
  booktitle={Proceedings of ICDE},
  pages={699--712},
  year={2022},
  organization={IEEE}
}

@inproceedings{ju2022kgnn,
  title={KGNN: Harnessing kernel-based networks for semi-supervised graph classification},
  author={Ju, Wei and Yang, Junwei and Qu, Meng and Song, Weiping and Shen, Jianhao and Zhang, Ming},
  booktitle={Proceedings of WSDM},
  pages={421--429},
  year={2022}
}

@article{luo2023towards,
  title={Towards semi-supervised universal graph classification},
  author={Luo, Xiao and Zhao, Yusheng and Qin, Yifang and Ju, Wei and Zhang, Ming},
  journal={IEEE TKDE},
  volume={36},
  number={1},
  pages={416--428},
  year={2023},
  publisher={IEEE}
}

@inproceedings{li2019semi,
  title={Semi-supervised graph classification: A hierarchical graph perspective},
  author={Li, Jia and Rong, Yu and Cheng, Hong and Meng, Helen and Huang, Wenbing and Huang, Junzhou},
  booktitle={Proceedings of the Web Conference},
  pages={972--982},
  year={2019}
}

@article{li2022semi,
  title={Semi-supervised hierarchical graph classification},
  author={Li, Jia and Huang, Yongfeng and Chang, Heng and Rong, Yu},
  journal={IEEE TPAMI},
  volume={45},
  number={5},
  pages={6265--6276},
  year={2022}
}

@inproceedings{you2020graph,
  title={Graph contrastive learning with augmentations},
  author={You, Yuning and Chen, Tianlong and Sui, Yongduo and Chen, Ting and Wang, Zhangyang and Shen, Yang},
  booktitle={Proceedings of NeurIPS},
  pages = {5812--5823},
  year={2020}
}

@inproceedings{you2021graph,
  title={Graph contrastive learning automated},
  author={You, Yuning and Chen, Tianlong and Shen, Yang and Wang, Zhangyang},
  booktitle={Proceedings of ICML},
  pages={12121--12132},
  year={2021},
  organization={PMLR}
}

@inproceedings{suninfograph,
  title={InfoGraph: Unsupervised and Semi-supervised Graph-Level Representation Learning via Mutual Information Maximization},
  author={Sun, Fan-Yun and Hoffman, Jordan and Verma, Vikas and Tang, Jian},
  booktitle={Proceedings of ICLR},
  pages={1-9},
  year={2020}
}

@inproceedings{ju2022tgnn,
  title={TGNN: A joint semi-supervised framework for graph-level classification},
  author={Ju, Wei and Luo, Xiao and Qu, Meng and Wang, Yifan and Chen, Chong and Deng, Minghua and Hua, Xian-Sheng and Zhang, Ming},
  booktitle={Proceedings of IJCAI},
  year={2022},
  pages={2122-2128}
}

@article{ju2022ghnn,
  title={GHNN: Graph harmonic neural networks for semi-supervised graph-level classification},
  author={Ju, Wei and Luo, Xiao and Ma, Zeyu and Yang, Junwei and Deng, Minghua and Zhang, Ming},
  journal={Neural Networks},
  volume={151},
  pages={70--79},
  year={2022},
  publisher={Elsevier}
}

@inproceedings{ju2024hypergraph,
  title={Hypergraph-enhanced Dual Semi-supervised Graph Classification},
  author={Ju, Wei and Mao, Zhengyang and Yi, Siyu and Qin, Yifang and Gu, Yiyang and Xiao, Zhiping and Wang, Yifan and Luo, Xiao and Zhang, Ming},
  booktitle={Proceedings of ICML},
  pages = 	 {22594--22604},
  year={2024}
}

@article{wu2024hierarchy,
  title={Hierarchy-aware adaptive graph neural network},
  author={Wu, Dengsheng and Wu, Huidong and Li, Jianping},
  journal={IEEE TKDE},
  volume={37},
  number={1},
  pages={365--378},
  year={2024},
  publisher={IEEE}
}

@article{lancichinetti2009detecting,
  title={Detecting the overlapping and hierarchical community structure in complex networks},
  author={Lancichinetti, Andrea and Fortunato, Santo and Kert{\'e}sz, J{\'a}nos},
  journal={New journal of physics},
  volume={11},
  number={3},
  pages={033015},
  year={2009},
  publisher={IOP Publishing}
}

@inproceedings{dasgupta2016cost,
  title={A cost function for similarity-based hierarchical clustering},
  author={Dasgupta, Sanjoy},
  booktitle={Proceedings of STOC},
  pages={118--127},
  year={2016}
}

@inproceedings{chami2019hyperbolic,
  title={Hyperbolic graph convolutional neural networks},
  author={Chami, Ines and Ying, Rex and Re, Christopher and Leskovec, Jure},
  booktitle={Proceedings of NeurIPS},
  pages={4868--4879},
  year={2019}
}

@inproceedings{bianchi2020spectral,
  title={Spectral clustering with graph neural networks for graph pooling},
  author={Bianchi, Filippo Maria and Grattarola, Daniele and Alippi, Cesare},
  booktitle={Proceedings of ICML},
  pages={874--883},
  year={2020},
  organization={PMLR}
}

@inproceedings{yang2024sebot,
  title={SeBot: Structural Entropy Guided Multi-View Contrastive Learning for Social Bot Detection},
  author={Yang, Yingguang and Wu, Qi and He, Buyun and Peng, Hao and Yang, Renyu and Hao, Zhifeng and Liao, Yong},
  booktitle={Proceedings of KDD},
  pages={3841--3852},
  year={2024}
}

@inproceedings{hao2020asgn,
  title={ASGN: An active semi-supervised graph neural network for molecular property prediction},
  author={Hao, Zhongkai and Lu, Chengqiang and Huang, Zhenya and Wang, Hao and Hu, Zheyuan and Liu, Qi and Chen, Enhong and Lee, Cheekong},
  booktitle={Proceedings of KDD},
  pages={731--752},
  year={2020}
}

@inproceedings{yue2022label,
  title={Label-invariant augmentation for semi-supervised graph classification},
  author={Yue, Han and Zhang, Chunhui and Zhang, Chuxu and Liu, Hongfu},
  booktitle={Proceedings of NeurIPS},
  pages={29350--29361},
  year={2022}
}

@article{li2016structural,
  title={Structural information and dynamical complexity of networks},
  author={Li, Angsheng and Pan, Yicheng},
  journal={IEEE TIT},
  volume={62},
  number={6},
  pages={3290--3339},
  year={2016}
}

@inproceedings{sunlsenet,
  title={LSEnet: Lorentz Structural Entropy Neural Network for Deep Graph Clustering},
  author={Sun, Li and Huang, Zhenhao and Peng, Hao and Wang, YuJie and Liu, Chunyang and Philip S., Yu},
  pages={47078 - 47104},
  year={2024},
  booktitle={Proceedings of ICML}
}

@inproceedings{zhou2006learning,
  title={Learning with hypergraphs: clustering, classification, and embedding},
  author={Zhou, Dengyong and Huang, Jiayuan and Sch{\"o}lkopf, Bernhard},
  booktitle={NeurIPS},
  pages={1601--1608},
  year={2006}
}

@article{shervashidze2011weisfeiler,
  title={Weisfeiler-lehman graph kernels.},
  author={Shervashidze, Nino and Schweitzer, Pascal and Van Leeuwen, Erik Jan and Mehlhorn, Kurt and Borgwardt, Karsten M},
  journal={JMLR},
  volume={12},
  number={9},
  year={2011}
}

@article{narayanan2017graph2vec,
  title={graph2vec: Learning distributed representations of graphs},
  author={Narayanan, Annamalai and Chandramohan, Mahinthan and Venkatesan, Rajasekar and Chen, Lihui and Liu, Yang and Jaiswal, Shantanu},
  journal={arXiv preprint arXiv:1707.05005},
  year={2017}
}

@inproceedings{adhikari2018sub2vec,
  title={Sub2vec: Feature learning for subgraphs},
  author={Adhikari, Bijaya and Zhang, Yao and Ramakrishnan, Naren and Prakash, B Aditya},
  booktitle={Proceedings of PAKDD},
  pages={170--182},
  year={2018},
}

@inproceedings{grandvalet2004semi,
  title={Semi-supervised learning by entropy minimization},
  author={Grandvalet, Yves and Bengio, Yoshua},
  booktitle={Proceedings of NeurIPS},
  pages={529--536},
  year={2004}
}

@inproceedings{tarvainen2017mean,
  title={Mean teachers are better role models: Weight-averaged consistency targets improve semi-supervised deep learning results},
  author={Tarvainen, Antti and Valpola, Harri},
  booktitle={Proceedings of NeurIPS},
  pages={1195--1204},
  year={2017}
}

@article{miyato2018virtual,
  title={Virtual adversarial training: a regularization method for supervised and semi-supervised learning},
  author={Miyato, Takeru and Maeda, Shin-ichi and Koyama, Masanori and Ishii, Shin},
  journal={IEEE TPAMI},
  volume={41},
  number={8},
  pages={1979--1993},
  year={2018}
}

@article{ju2024focus,
  title={Focus on informative graphs! Semi-supervised active learning for graph-level classification},
  author={Ju, Wei and Mao, Zhengyang and Qiao, Ziyue and Qin, Yifang and Yi, Siyu and Xiao, Zhiping and Luo, Xiao and Fu, Yanjie and Zhang, Ming},
  journal={Pattern Recognition},
  volume={153},
  pages={110567},
  year={2024},
  publisher={Elsevier}
}

@inproceedings{kim2024survey,
  title={A survey on hypergraph neural networks: An in-depth and step-by-step guide},
  author={Kim, Sunwoo and Lee, Soo Yong and Gao, Yue and Antelmi, Alessia and Polato, Mirko and Shin, Kijung},
  booktitle={Proceedings of KDD},
  pages={6534--6544},
  year={2024}
}

@article{chodrow2021generative,
  title={Generative hypergraph clustering: From blockmodels to modularity},
  author={Chodrow, Philip S and Veldt, Nate and Benson, Austin R},
  journal={Science Advances},
  volume={7},
  number={28},
  pages={eabh1303},
  year={2021},
}

@inproceedings{feng2019hypergraph,
  title={Hypergraph neural networks},
  author={Feng, Yifan and You, Haoxuan and Zhang, Zizhao and Ji, Rongrong and Gao, Yue},
  booktitle={Proceedings of AAAI},
  pages={3558--3565},
  year={2019}
}

@article{gao2022hgnn+,
  title={HGNN+: General hypergraph neural networks},
  author={Gao, Yue and Feng, Yifan and Ji, Shuyi and Ji, Rongrong},
  journal={IEEE TPAMI},
  volume={45},
  number={3},
  pages={3181--3199},
  year={2022},
  publisher={IEEE}
}

@article{sun2023self,
  title={Self-supervised hypergraph representation learning for sociological analysis},
  author={Sun, Xiangguo and Cheng, Hong and Liu, Bo and Li, Jia and Chen, Hongyang and Xu, Guandong and Yin, Hongzhi},
  journal={IEEE TKDE},
  volume={35},
  number={11},
  pages={11860--11871},
  year={2023},
}

@inproceedings{grover2016node2vec,
  title={node2vec: Scalable feature learning for networks},
  author={Grover, Aditya and Leskovec, Jure},
  booktitle={Proceedings of KDD},
  pages={855--864},
  year={2016}
}

@inproceedings{kipf2017semi,
  title={Semi-Supervised Classification with Graph Convolutional Networks},
  author={Kipf, Thomas N and Welling, Max},
  booktitle={Proceedings of ICLR},
  year={2017},
  pages={1-14}
}

@inproceedings{wu2022structural,
  title={Structural entropy guided graph hierarchical pooling},
  author={Wu, Junran and Chen, Xueyuan and Xu, Ke and Li, Shangzhe},
  booktitle={Proceedings of ICML},
  pages={24017--24030},
  year={2022},
}

@article{ren2024hi,
  title={Hi-PART: Going Beyond Graph Pooling with Hierarchical Partition Tree for Graph-Level Representation Learning},
  author={Ren, Yuyang and Zhang, Haonan and Fu, Luoyi and Liang, Shiyu and Zhou, Lei and Wang, Xinbing and Cao, Xinde and Long, Fei and Zhou, Chenghu},
  journal={ACM TKDD},
  volume={18},
  number={4},
  pages={1--20},
  year={2024},
}

@inproceedings{morris2020tudataset,
    title={{TUDataset}: A collection of benchmark datasets for learning with graphs},
    author={Morris, Christopher and Kriege, Nils M. and Bause, Franka and Kersting, Kristian and Mutzel, Petra and Neumann, Marion},
    booktitle={ICML 2020 Workshop on Graph Representation Learning and Beyond (GRL+)},
    year={2020},
    url={https://arxiv.org/abs/2007.08663}
}

@article{borgwardt2005protein,
  title={Protein function prediction via graph kernels},
  author={Borgwardt, Karsten M and Ong, Cheng Soon and Sch{\"o}nauer, Stefan and Vishwanathan, SVN and Smola, Alex J and Kriegel, Hans-Peter},
  journal={Bioinformatics},
  volume={21},
  number={suppl\_1},
  pages={i47--i56},
  year={2005},
}

@article{dobson2003distinguishing,
  title={Distinguishing enzyme structures from non-enzymes without alignments},
  author={Dobson, Paul D and Doig, Andrew J},
  journal={Journal of molecular biology},
  volume={330},
  number={4},
  pages={771--783},
  year={2003},
  publisher={Elsevier}
}

@inproceedings{yanardag2015deep,
  title={Deep graph kernels},
  author={Yanardag, Pinar and Vishwanathan, SVN},
  booktitle={Proceedings of the KDD},
  pages={1365--1374},
  year={2015}
}

@article{neumann2016propagation,
  title={Propagation kernels: efficient graph kernels from propagated information},
  author={Neumann, Marion and Garnett, Roman and Bauckhage, Christian and Kersting, Kristian},
  journal={Machine learning},
  volume={102},
  pages={209--245},
  year={2016},
  publisher={Springer}
}

@inproceedings{takai2020hypergraph,
  title={Hypergraph clustering based on pagerank},
  author={Takai, Yuuki and Miyauchi, Atsushi and Ikeda, Masahiro and Yoshida, Yuichi},
  booktitle={Proceedings of KDD},
  pages={1970--1978},
  year={2020}
}

@book{li2024science,
  author    = {Angsheng Li},
  title     = {Science of Artificial Intelligence: Mathematical Principles of Intelligence (In Chinese)},
  publisher = {Science Press},
  year      = {2024},
  address   = {Beijing}
}

@inproceedings{zeng2026structural,
  title={Structural Information-based Hierarchical Diffusion for Offline Reinforcement Learning},
  author={Zeng, Xianghua and Peng, Hao and Pan, Yicheng and Li, Angsheng and Wu, Guanlin},
  booktitle={Proceedings of NeurIPS},
  pages={109539--109567},
  year={2026}
}

@inproceedings{zhu2023hitin,
  title={Hitin: Hierarchy-aware tree isomorphism network for hierarchical text classification},
  author={Zhu, He and Zhang, Chong and Huang, Junjie and Wu, Junran and Xu, Ke},
  booktitle={Proceedings of ACL},
  pages={7809--7821},
  year={2023}
}

@inproceedings{wilder2019end,
  title={End to end learning and optimization on graphs},
  author={Wilder, Bryan and Ewing, Eric and Dilkina, Bistra and Tambe, Milind},
  booktitle={Proceedings of NeurIPS},
  year={2019},
  pages={1-9}
}

@article{wang2025disensemi,
  title={DisenSemi: Semi-supervised graph classification via disentangled representation learning},
  author={Wang, Yifan and Luo, Xiao and Chen, Chong and Hua, Xian-Sheng and Zhang, Ming and Ju, Wei},
  journal={IEEE TNNLS},
  volume={36},
  number={5},
  pages={8192--8204},
  year={2025},
}

@article{zhang2025enhanced,
  title={Enhanced pre-training for recommendation via hypergraph structural entropy},
  author={Zhang, Jingyun and Peng, Hao and Yang, Mingdai and Yu, Philip S},
  journal={ACM TOIS},
  volume={44},
  number={2},
  pages={1--48},
  year={2025},
}

@inproceedings{zeng2026hyperbolic,
  title={Hyperbolic continuous structural entropy for hierarchical clustering},
  author={Zeng, Guangjie and Peng, Hao and Li, Angsheng and Sun, Li and Liu, Chunyang and Li, Shengze and Pan, Yicheng and Yu, Philip S},
  booktitle={Proceedings of AAAI},
  volume={40},
  number={33},
  pages={28094--28102},
  year={2026}
}

@article{huang2026sector,
  title={SECTOR: structural entropy-based learning of spatiotemporal organisation in spatial transcriptomics},
  author={Huang, Li and Zhang, Jingyun and Gong, Weikang and Zeng, Guangjie and Peng, Hao and Chen, Dongsheng},
  journal={Bioinformatics},
  volume={42},
  number={6},
  pages={btag367},
  year={2026},
}

@article{li2023dedoc2,
  title={DeDoc2 Identifies and Characterizes the Hierarchy and Dynamics of Chromatin TAD-Like Domains in the Single Cells},
  author={Li, Angsheng and Zeng, Guangjie and Wang, Haoyu and Li, Xiao and Zhang, Zhihua},
  journal={Advanced Science},
  volume={10},
  number={20},
  pages={2300366},
  year={2023},
}

@inproceedings{liu2026effective,
  title={Effective and Unsupervised Social Event Detection and Evolution via RAG and Structural Entropy},
  author={Liu, Qitong and Peng, Hao and Li, Zuchen and Meng, Xihang and Yang, Ziyu and Li, Jiting and Sun, Li and Yu, Philip S},
  booktitle={Proceedings of WWW},
  pages={4529--4540},
  year={2026}
}

@ARTICLE{meng2026DHMNN,
  author={Meng, Xihang and Peng, Hao and Zeng, Guangjie and Sun, Li and Hao, Zhifeng and Yu, Philip S.},
  journal={TNNLS}, 
  title={DHMNN: A Hypergraph Motif-Based Framework for Directed Hyperlink Prediction}, 
  year={2026},
  volume={},
  number={},
  pages={1-15},
}

\clearpage

\appendix

\section{Proofs}
\subsection{Proof of Theorem 3.1}\label{sec:theorem1}
\setcounter{theorem}{0}
\begin{theorem}[Connection to Classic Structural Entropy]\label{theorem:degenerate_sup}
The HSE of hypergraph $H$ transformed from a simple graph $G$ w.r.t. any given partitioning tree $\mathcal{T}$ equals to $\frac{1}{2}\mathcal{H}^\mathcal{T}(G)$ in Equation (2).
\end{theorem}

\noindent\emph{Proof.}
Let $H=(\mathcal{V}, \mathcal{E}_H, \mathbf{W})$ along with its incidence matrix $\mathbf{H}$ be the equivalent hypergraph of the simple undirected weighted graph $G=(\mathcal{V}, \mathcal{E}, \mathcal{W})$ by taking edges in $\mathcal{E}$ as hyperedges in $\mathcal{E}_H$, where $\mathcal{W} \in \mathbb{R}^{|\mathcal{V}| \times |\mathcal{V}|}$ is the edge weights matrix of $G$.
We have $\mathcal{E}_H = \mathcal{E}$ and $\mathbf{W} = \mathbf{I}_{|\mathcal{E}_H|}$, where $\mathbf{I}_{|\mathcal{E}_H|}$ is an identity matrix.
For an edge $\epsilon = (i,j)$, a hyperedge $e_H \in \mathcal{E}_H$ exists and $\mathbf{H}_{i\epsilon} = \mathbf{H}_{j\epsilon} = \mathcal{W}_{ij}$ is the edge weight.
Then, the hyperedge weight $\mathbf{W}_{\epsilon\epsilon} = 1$ and the hyperedge degree $\mathbf{B}_{\epsilon \epsilon} = \sum_{i=1}^{|\mathcal{V}|} \mathbf{H}_{i\epsilon} = 2 \mathcal{W}_{ij}$.
According to Equation (8), the hypergraph cut of $k$-th tree node at depth $h$ named $\alpha_k^h$ is:
\begin{equation}
\begin{aligned}
\mathbf{M}_k^h 
&= \sum_{\epsilon=1}^{|\mathcal{E}_H|} 
\left[ \frac{\mathbf{W}_{\epsilon \epsilon}}{\mathbf{B}_{\epsilon \epsilon}}
   \sum_{i=1}^{N}  \mathbf{S}_{ik}^h \mathbf{H}_{i\epsilon} 
    \sum_{j=1}^N  \mathbf{H}_{j\epsilon} (1-\mathbf{S}_{jk}^h) \right] \\
&= \sum_{\epsilon=1}^{|\mathcal{E}_H|} \left[\frac{1}{2\mathcal{W}_{ij}}
\sum_{i=1}^{N} \mathbb{I}(i\in T_{\alpha_k^h})\mathbf{H}_{i\epsilon}
\sum_{j=1}^{N} \mathbb{I}(j\notin T_{\alpha_k^h})\mathbf{H}_{j\epsilon}\right], \\
\end{aligned}
\end{equation}
where $\mathbf{S}_{ik}^h = \mathbb{I}(i\in T_{\alpha_k^h})$ indicates whether vertex $i$ belongs to the associated vertex set $T_{\alpha_k^h}$.
Since each hyperedge $\epsilon$ is transformed from a simple edge containing two vertices, it contributes to the hypergraph cut when only one of its vertices belongs to $T_{\alpha_k^h}$.
We can then rewrite the hypergraph cut as follows:
\begin{equation}
\begin{aligned}
\mathbf{M}_k^h 
&= \sum_{\epsilon=1}^{|\mathcal{E}_H|} \left[\frac{1}{2\mathcal{W}_{ij}}
\sum_{i=1}^{N}\sum_{j\in \mathcal{N}(i)}\mathbb{I}(i\in T_{\alpha_k^h})\mathbb{I}(j\notin T_{\alpha_k^h}) (\mathcal{W}_{ij})^2\right] \\
&= \frac{1}{2} \sum_{\epsilon=1}^{|\mathcal{E}_H|} \left[
\sum_{i=1}^{N}\sum_{j\in \mathcal{N}(i)}\mathbb{I}(i\in T_{\alpha_k^h})\mathbb{I}(j\notin T_{\alpha_k^h}) \mathcal{W}_{ij}\right] \\
&= \frac{1}{2} g_{\alpha_h^k},
\end{aligned}
\end{equation}
where $g_{\alpha_h^k}$ is the graph cut of tree node $\alpha_h^k$ w.r.t. $G$.
According to Equation (7), the hypergraph volumes of $k$-th tree node at depth $h$ named $\alpha_k^h$ is defined as follows:
\begin{equation}
\begin{aligned}
\mathbf{V}_k^h &= \sum_{i=1}^N \mathbf{S}_{ik}^h \mathbf{D}_{ii} 
= \sum_{i=1}^N \mathbb{I}(i \in T_{\alpha_k^h}) \sum_{\epsilon=1}^{|\mathcal{E}_H|} \mathbf{W}_{\epsilon\epsilon}\mathbf{H}_{i\epsilon} \\
&= \sum_{i=1}^N \mathbb{I}(i \in T_{\alpha_k^h}) \sum_{j=1}^N \mathcal{W}_{ij} 
= V_{\alpha_h^k},
\end{aligned}
\end{equation}
where $V_{\alpha_h^k}$ is the graph volume of tree node $\alpha_h^k$ w.r.t. $G$.
According to the level-wise assignment matrix in Definition 3 by \cite{sunlsenet}, the ascendant-descendant relationships between tree nodes from $h$-th to $h-1$-th level can be represented by an assignment matrix $\mathbf{C}^h$.
The ancestor volume of tree node $\alpha_h^k$ can be formulated by the assignment matrix $\mathbf{C}^h$ as follows:
\begin{equation}
\mathbf{V}^{h-1}_{k^-} = \sum_{k'=1}^{N_{h-1}}\mathbf{C}_{kk^\prime}^h V^{h-1}_{k'}
= V_{\alpha_{k^-}^{h-1}},
\end{equation}
where tree node $\alpha_{k^-}^{h-1}$ is the parent of node $\alpha_{k}^{h}$ and $V_{\alpha_{k^-}^{h-1}}$ is the volume of node $\alpha_{k^-}^{h-1}$.

Taken together, the HSE of hypergraph $H$ transformed from a simple graph $G$ in Definition 3.1 can be rewritten as follows:
\begin{equation}
\begin{aligned}
\mathcal{H}^\mathcal{T}(H) &= \sum_{h=1}^K \mathcal{H}^\mathcal{T}(H;h) = \sum_{h=1}^K - \frac{1}{V_H}  \sum_{k=1}^{N_h} \mathbf{M}_k^h \log_2 \frac{\mathbf{V}_k^h}{\mathbf{V}_{k^-}^{h-1}} \\ 
&= \sum_{h=1}^K - \frac{1}{V_G}  \sum_{k=1}^{N_h} \frac{1}{2} g_{\alpha_k^h} \log_2 \frac{V_{\alpha_h^k}}{V_{\alpha_{h-1}^{k^-}}} \\
&= \frac{1}{2}\sum_{\alpha\in\mathcal{T},\alpha\neq\lambda} - \frac{g_\alpha}{V_G}\log_2\frac{V_\alpha}{V_{\alpha^-}} \\
&= \frac{1}{2}\mathcal{H}^\mathcal{T}(G),
\end{aligned}
\end{equation}
which completes the proof.

\subsection{Proof of Theorem 3.2}\label{sec:theorem2}
We first introduce the Additive property of the one-dimensional HSE of hypergraph $H$, i.e., the HSE of the hypergraph when depth $K=1$, in Lemma 1 following \cite{sunlsenet}.
We then complete the proof of Theorem 3.2 using the Additive property.
\setcounter{theorem}{0}
\begin{lemma}[Additive] The one-dimensional hypergraph structural entropy of a hypergraph $H$ has the following property:
\begin{equation} \label{eq:additive}
\mathcal{H}^1(H) \leq \sum_{h=1}^K \sum_{j=1}^{N_{h-1}} \frac{\mathbf{V}_j^{h-1}}{V_H}\mathbb{E} ( [ \frac{\mathbf{C}_{kj}^h \mathbf{V}_k^h}{\mathbf{V}_j^{h-1}} ]_{k=1,\ldots,N_h} ),
\end{equation}
where $\mathbb{E}(p_1,\ldots,p_n) = - \sum_{i=1}^n p_i \log_2 p_i$ is the Shannon entropy.
\end{lemma}
\noindent\emph{Proof.}
On the one hand, the one-dimensional HSE can be represented as follows:
\begin{equation}
\begin{aligned}
\mathcal{H}^1(H) &= - \frac{1}{V_H}\sum_{i=1}^{N}\mathbf{M}_i^1 \log_2 \frac{\mathbf{V}_i^1}{\mathbf{V}_{i^-}^{0}}.
\end{aligned}
\end{equation}
Since the corresponding partitioning tree of one-dimensional HSE is at depth $1$, the hypergraph volume of each tree leaf is the vertex degree $\mathbf{D}_{ii}$, and the hypergraph volume of its parent is $V_H$.
Since $\mathbf{S}^1$ is an identity matrix $\mathbf{I}_{N}$, the hypergraph cut term $\mathbf{M}_i^1$ can be expanded as follows:
\begin{equation}
\begin{aligned}
\mathbf{M}^1_k &= \sum_{\epsilon=1}^{|\mathcal{E}_H|} \left[ \frac{\mathbf{W}_{\epsilon\epsilon}}{\sum_{j=1}^N \mathbf{H}_{j\epsilon}} \mathbf{H}_{k\epsilon} \sum_{j=1,j \neq k}^N \mathbf{H}_{j\epsilon} \right] \\
&\leq \sum_{\epsilon=1}^{|\mathcal{E}_H|} \left[ \frac{\mathbf{W}_{\epsilon\epsilon}}{\sum_{j=1}^N \mathbf{H}_{j\epsilon}} \mathbf{H}_{k\epsilon} \sum_{j=1}^N \mathbf{H}_{j\epsilon} \right] \\
&= \sum_{\epsilon=1}^{|\mathcal{E}_H|} \left[ \mathbf{W}_{\epsilon\epsilon} \mathbf{H}_{k\epsilon} \right] \\
&= \mathbf{D}_{kk}.
\end{aligned}
\end{equation}
Then, we can derive the following property about one-dimensional HSE:
\begin{equation}
\mathcal{H}^1(H) \leq - \frac{1}{V_H} \sum_{k=1}^N \mathbf{D}_{kk} \log_2 \frac{\mathbf{D}_{kk}}{V_H}.
\end{equation}

On the other hand, the right-hand side of Equation (\ref{eq:additive}) can be expanded as follows:
\begin{equation}
\begin{aligned}
\mathcal{H}^{\mathbb{E}}(H) &= - \sum_{h=1}^K\sum_{j=1}^{N_{h-1}}\frac{\mathbf{V}_j^{h-1}}{V_H}\sum_{k=1}^{N_h}\frac{\mathbf{C}_{kj}^h\mathbf{V}_k^h}{\mathbf{V}_j^{h-1}}\log_2 \frac{\mathbf{C}_{kj}^h\mathbf{V}_k^h}{\mathbf{V}_j^{h-1}} \\
&= - \frac{1}{V_H}\sum_{h=1}^K \sum_{k=1}^{N_h}\sum_{j=1}^{N_{h-1}} \mathbf{C}_{kj}^h\mathbf{V}_k^h \log_2 \frac{\mathbf{C}_{kj}^h \mathbf{V}_k^h}{\mathbf{V}_j^{h-1}} \\
&= - \frac{1}{V_H}\sum_{h=1}^K \sum_{k=1}^{N_h} \mathbf{V}_k^h \log_2 \frac{\mathbf{V}_k^h}{\mathbf{V}_{k^-}^{h-1}},
\end{aligned}
\end{equation}
where $\mathbf{V}_{k^-}^{h-1}$ is the volume of tree node $\alpha_{k^-}^{h-1}$, which is the parent node of tree node $\alpha_{k}^{h}$. Since when and only when $\mathbf{C}_{kj}^h = 1$, tree node $\alpha_{j}^{h-1}$ is the parent of node $\alpha_{k}^{h}$, we omit some $j$ when $\mathbf{C}_{kj}^{h} = 0$.

Then, let us subtract the right-hand side from the left-hand side of Equation (\ref{eq:additive}) as follows:
\begin{equation}
\begin{aligned}
A &= \mathcal{H}^1(H) - \mathcal{H}^{\mathbb{E}}(H) \\
&= \mathcal{H}^1(H) - \sum_{h=1}^K \sum_{j=1}^{N_{h-1}} \frac{\mathbf{V}_j^{h-1}}{V_H}\mathbb{E} ( [ \frac{\mathbf{C}_{kj}^h \mathbf{V}_k^h}{\mathbf{V}_j^{h-1}} ]_{k=1,\ldots,N_h} ) \\
&\leq  - \frac{1}{V_H} \sum_{k=1}^N \mathbf{D}_{kk} \log_2 \frac{\mathbf{D}_{kk}}{V_H} + \frac{1}{V_H}\sum_{h=1}^K \sum_{k=1}^{N_h} \mathbf{V}_k^h \log_2 \frac{\mathbf{V}_k^h}{\mathbf{V}_{k^-}^{h-1}} \\ 
&= - \frac{1}{V_H} \sum_{k=1}^N \mathbf{D}_{kk} \log_2 \frac{\mathbf{D}_{kk}}{V_H} + \frac{1}{V_H}\sum_{h=1}^K\sum_{k=1}^{N_h}\mathbf{V}_k^h \log_2 \frac{\mathbf{V}_k^h}{V_H} \\ &\ \ \ \  - \frac{1}{V_H}\sum_{h=1}^K\sum_{k=1}^{N_h}\mathbf{V}_k^h \log_2 \frac{\mathbf{V}_{k^-}^{h-1}}{V_H}. \\
\end{aligned}
\end{equation}
When $h=K$, we have $\mathbf{V}_k^h = \mathbf{D}_{kk}$.
Thus, the process continues as:
\begin{equation}
\begin{aligned}
A &\leq \frac{1}{V_H} \left[ \sum_{h=1}^{K-1}\sum_{k=1}^{N_h}\mathbf{V}_k^h \log_2 \frac{\mathbf{V}_k^h}{V_H} - \sum_{h=2}^K \sum_{k=1}^{N_h} \mathbf{V}_k^h \log_2 \frac{\mathbf{V}_{k^-}^{h-1}}{V_H} \right] \\ 
&= \frac{1}{V_H} \sum_{h=2}^K \left[ \sum_{j=1}^{N_{h-1}}\mathbf{V}_j^{h-1} \log_2 \frac{\mathbf{V}_j^{h-1}}{V_H} - \sum_{k=1}^{N_h} \mathbf{V}_k^h \log_2 \frac{\mathbf{V}_{k^-}^{h-1}}{V_H} \right] \\ 
&=  \sum_{h=2}^K \left[ \sum_{j=1}^{N_{h-1}}\sum_{k=1}^{N_h} \frac{\mathbf{C}_{kj}^h\mathbf{V}_k^h}{V_H} \log_2 \frac{\mathbf{V}_j^{h-1}}{V_H} -  \sum_{k=1}^{N_h} \frac{\mathbf{V}_k^h}{V_H} \log_2 \frac{\mathbf{V}_{k^-}^{h-1}}{V_H} \right] \\ 
&= \sum_{h=2}^K \left[ \sum_{k=1}^{N_h} \frac{\mathbf{V}_k^h}{V_H} \log_2 \frac{\mathbf{V}_{k^-}^{h-1}}{V_H} -  \sum_{k=1}^{N_h} \frac{\mathbf{V}_k^h}{V_H} \log_2 \frac{\mathbf{V}_{k^-}^{h-1}}{V_H} \right] \\ 
&= 0.
\end{aligned}
\end{equation}
Taken together, we have 
\begin{equation}
\begin{aligned}
A &= \mathcal{H}^1(H) - \mathcal{H}^{\mathbb{E}}(H) \\
&= \mathcal{H}^1(H) - \sum_{h=1}^K \sum_{j=1}^{N_{h-1}} \frac{\mathbf{V}_j^{h-1}}{V_H}\mathbb{E} ( [ \frac{\mathbf{C}_{kj}^h \mathbf{V}_k^h}{\mathbf{V}_j^{h-1}} ]_{k=1,\ldots,N_h} ) \leq 0,
\end{aligned}
\end{equation}
which completes the proof.

\setcounter{theorem}{1}
\begin{theorem}[Connection to Hypergraph Clustering]\label{theorem:clustering_sup}
Given a hypergraph $H=(\mathcal{V}, \mathcal{E}_H, \mathbf{W})$ along with its incidence matrix $\mathbf{H}$, the normalized HSE of $H$ is defined as $\rho(H) = \mathcal{H}^K(H) / \mathcal{H}^1(H)$, and $\Phi(H) = \min_{S \subseteq \mathcal{V}} \{ \frac{V_{\partial S}}{\min \{ V_{S}, V_{S^c} \}}\}$ is the hypergraph conductance. The following inequality holds:
\begin{equation}
\rho^K(H) \geq \Phi(H).
\end{equation}
\end{theorem}

\noindent\emph{Proof.}
For each tree node $\alpha_k^h$, its associated vertex set is $T_{\alpha_k^h}$ and its hypergraph cut is $V_{\partial T_{\alpha_k^h}}=\mathbf{M}_k^h$, its hypergraph conductance is:
\begin{equation}
\phi_k^h = \frac{V_{\partial T_{\alpha_k^h}}}{\min \{ V_{T_{\alpha_k^h}}, V_H - V_{T_{\alpha_k^h}} \}} = \frac{\mathbf{M}_k^h}{\min \{ \mathbf{V}_k^h, V_H -\mathbf{V}_k^h \}}.
\end{equation}
Without loss of generality, we assume that $\mathbf{V}_k^h \leq \frac{1}{2}V_H$ such that $\min \{ \mathbf{V}_k^h, V_H - \mathbf{V}_k^h \} = \mathbf{V}_k^h$.
The hypergraph conductance of $H$ is:
\begin{equation}
\Phi(H)=\min_{S\subseteq \mathcal{V}} \Phi(S) = \min_{S \subseteq \mathcal{V}} \{ \frac{V_{\partial S}}{\min \{ V_{S}, V_{S^c} \}}\},
\end{equation}
where $S$ traverses all possible vertex subsets of $\mathcal{V}$.
According to Definition 4, the HSE of hypergraph $H$ with a partitioning tree $\mathcal{T}$ at depth $K$ can be formulated as follows:
\begin{equation}
\begin{aligned}
\mathcal{H}^\mathcal{T}(H) &= - \frac{1}{V_H}\sum_{h=1}^K\sum_{k=1}^{N_h}\mathbf{M}_k^h \log_2 \frac{\mathbf{V}_k^h}{\mathbf{V}_{k^-}^{h-1}} \\ 
&= - \frac{1}{V_H} \sum_{h=1}^K\sum_{k=1}^{N_h} \phi_h^k \mathbf{V}_k^h \log_2 \frac{\mathbf{V}_k^h}{\mathbf{V}_{k^-}^{h-1}}, \\
\end{aligned}
\end{equation}
Since $\Phi(H) \leq \phi_k^h$ for all tree node $\alpha_k^h$, we have:
\begin{equation}
\begin{aligned}
\mathcal{H}^\mathcal{T}(H) &\geq -\Phi(H)\sum_{h=1}^{K}\sum_{k=1}^{N_h}\frac{\mathbf{V}_k^h}{V_H} \log_2 \frac{\mathbf{V}_k^h}{\mathbf{V}_{k^-}^{h-1}} \\
&= - \Phi(H) \sum_{h=1}^K \sum_{k=1}^{N_h} \frac{\sum_{j=1}^{N_{h-1}}\mathbf{C}_{kj}^h \mathbf{V}_{k}^h}{V_H} \log_2 \frac{\mathbf{V}_k^h}{\mathbf{V}_{k^-}^{h-1}} \\ 
&= - \Phi(H) \sum_{h=1}^K \sum_{j=1}^{N_{h-1}} \frac{\mathbf{V}_j^{h-1}}{V_H} \sum_{k=1}^{N_h} \frac{\mathbf{C}_{kj}^h \mathbf{V}_{k}^h}{\mathbf{V}_j^{h-1}} \log_2 \frac{\mathbf{V}_k^h}{\mathbf{V}_{k^-}^{h-1}} \\ 
&= - \Phi(H) \sum_{h=1}^K \sum_{j=1}^{N_{h-1}} \frac{\mathbf{V}_j^{h-1}}{V_H} \sum_{k=1}^{N_h} \frac{\mathbf{C}_{kj}^h \mathbf{V}_{k}^h}{\mathbf{V}_j^{h-1}} \\ &\ \ \ \ \cdot \log_2 \frac{\mathbf{V}_k^h}{\sum_{m=1}^{N_{h-1}}\mathbf{C}_{km}^h\mathbf{V}_m^{h-1}} \\ 
&= - \Phi(H) \sum_{h=1}^K \sum_{j=1}^{N_{h-1}} \frac{\mathbf{V}_j^{h-1}}{V_H} \sum_{k=1}^{N_h} \frac{\mathbf{C}_{kj}^h \mathbf{V}_{k}^h}{\mathbf{V}_j^{h-1}} \log_2 \frac{\mathbf{C}_{kj}^h\mathbf{V}_k^h}{\mathbf{V}_j^{h-1}} \\ 
&= \Phi(H) \sum_{h=1}^K \sum_{j=1}^{N_{h-1}} \frac{\mathbf{V}_j^{h-1}}{V_H} \mathbb{E} ([\frac{\mathbf{C}_{kj}^h\mathbf{V}_k^h}{\mathbf{V}_j^{h-1}}]_{k=1,\ldots,N_h}) \\ 
&\geq \Phi(H) \mathcal{H}^1(H). 
\end{aligned}
\end{equation}
Here, we use the inverse trick about $\mathbf{C}_{kj}^h$ mentioned in Appendix \ref{sec:theorem2}. Since $\mathcal{H}^\mathcal{T}(H)\geq \Phi(H) \mathcal{H}^1(H) > 0$, we have:
\begin{equation}
\rho^\mathcal{T}(H) = \frac{\mathcal{H}^\mathcal{T}(H)}{\mathcal{H}^1(H)} \geq \Phi(H).
\end{equation}
Since the above inequality holds for every $K$-dimensional partitioning tree $\mathcal{T}$ of hypergraph $H$, $\rho^K (H) \geq \Phi(H)$ holds.

\end{document}